\documentclass[review]{elsarticle}

\usepackage{amssymb}
\usepackage{graphicx}
\usepackage{dcolumn}% Align table columns on decimal point
\usepackage{bm}% bold math

\usepackage{eqnarray}
\usepackage{amsmath}
\usepackage{amsfonts}
\usepackage{hyperref}
\usepackage{booktabs}
\usepackage{graphicx, subfigure}
\usepackage{float}
\usepackage{relsize}
\usepackage{array}
\usepackage{xcolor}
\usepackage{upgreek}
\journal{Computational Materials Science}

\begin{document}
%\tableofcontents
\begin{frontmatter}

%\title{Visible fingerprint of X-ray images of epoxy resins using  singular value decomposition of  deep learning  features}

\title{Visualizing key features in X-ray images of epoxy resins for improved material classification using singular value decomposition of deep learning features}

%Eigenfeatures: Discrimination of x-ray images of epoxy resins using singular value decomposition of deep learning features

%% Group authors per affiliation:

\author{Edgar Avalos, Kazuto Akagi and Yasumasa Nishiura$^1$}
\address{Mathematical Science Group, WPI-Advanced Institute for Materials Research (AIMR), Tohoku University, Japan\\
$^1$Research Institute for Electronic Science, Hokkaido University and MathAM-OIL, Tohoku University and AIST, Japan
}

\begin{abstract}
Although  the process variables of epoxy resins alter their mechanical properties, recently
it was found that
the total variation of the X-ray images of  these resins is one of the key features
that affect the toughness of these materials. However it is still not
clear how to visualize such a difference in a clear way.
%
 %the visual identification of the characteristic  features of  X-ray images of samples of these materials is challenging.
%
%
To facilitate the visualization, we use a robust approximation of the gradient of the intensity field of  the X-ray images of different kinds of epoxy resins and then we use deep learning to discover the most representative features of the transformed images.
In  this solution of the inverse problem to find characteristic features to discriminate samples of heterogeneous materials, we use the eigenvectors obtained from the singular value decomposition of  all the channels of the  response maps of the early layers in a convolutional neural network.
While the strongest activated channel gives a visual representation of the characteristic features, often these are not robust enough in some  practical settings. On the other hand, the left singular vectors of  the matrix decomposition of the response maps barely change when variables such as the capacity of the network or the network architecture change.
High classification accuracy and robustness of characteristic features are presented in this work.
\footnote{This article appears in Computational Materials Science
Volume 186, January 2021, 109996. \href{url}{https://doi.org/10.1016/j.commatsci.2020.109996}}
\end{abstract}

 
\end{frontmatter}
%\linenumbers

%%%%%%
 \section{Introduction}

 X-ray CT imaging of polymer composites enables the non-destructive visualization of three-dimensional samples as well as two-dimensional slices of the material~\cite{Garcea2018}. Despite the significative improvements in the spatial resolution, the resulting images exhibit highly fluctuating electron density patterns that are extremely difficult to discriminate by simple observation. 
 %
 % May 10th add per Nishiura request:
These density patterns describe the  structural heterogeneity of the epoxy resins, and the heterogeneity at the micro/meso-level has a profound relation with the macroscopic performance~\cite{torquato2013random,cross,cross2}. Therefore the visual identification  of characteristic features in the X-ray images is desirable to understand the mechanical behavior of these materials.
 %+++++
 % end May 10th
%
%The problem of grouping images of samples of materials with similar properties was recently addressed in Ref.~\cite{Avalos2020}. 
%However, while it is possible to categorize X-ray images with similar properties, the associated inverse problem of  finding out what are the most relevant features in the images that differentiate samples from one another, remains unsolved. In this paper we propose a solution  to the inverse problem based on deep learning.
%

% May 10th add per Nishiura request:
The problem of grouping images of samples of materials with similar properties was recently addressed in Ref.~\cite{Avalos2020},
where the total variation (TV) of an X-ray image defined as $\int \sqrt{f_x^2+f_y^2}~ dxdy$, is used to order the images into groups with similar mechanical performance. 
%, in which the sample No. 3 is the best performing material among four samples.
In the expression for the TV, $f_x$ and $f_y$  are the gradients of the intensity field along the $x$ and $y$ directions, respectively.
Remarkably, the aforementioned work presents the TV as a property that describes the performance of materials,  which is a significant step towards the solution of the inverse problem.
However, while it is possible to use the total gradient to categorize X-ray images with similar properties, 
what is still lacking in the description of the amorphous materials is a
visual  fingerprint   of the most representative 
features that can be used both for discrimination and for describing  the performance of materials. 
A notable aspect in this work is that we
succeed in providing with visual qualitative representation of key features with the aid
of a neural network and singular value decomposition.
In this paper  we propose a solution to the inverse problem based on deep learning.
%
 %+++++
%Of course the visualized key feature is proportional to
%the strength of the quantity of TV so that the result is consistent with
%that of [2].
%(I did not check (don't know) how we can measure the strength of TV is
%proportional to that of the eigenfeature, but we can claim at least qualitatively
%the visible fingerprint is related to the amount of TV.)
% %+++++
 % end May 10th

Machine learning  tools  are  widely used to identify and categorize different samples of materials and to predict their properties~\cite{Petrich2017,Frankel2019,Hwang2019,Schmidt2019,Schwarzer2019}.
Some examples of these tools include deep convolutional neural networks (CNN)~\cite{LeCun1989,Krizhevsky2012,LeCun2015,Song2020}, which are highly accurate to classify images by minimizing  a suitable cost function~\cite{Bishop1995}.
By visualizing  the intermediate responses in a CNN one can have a better understanding of the features that the net uses for classification~\cite{Zeiler2014,Chollet2018}.
Among some few applications that take advantage of this approach, we can mention Ref.~\cite{Ziletti2018}, in which the authors use a CNN that is able not only to  classify crystals with astonishing accuracy, but remarkably  the activated filters of the early layers produce visual fingerprints that distinguish among different crystal symmetries. 
Similarly in Ref.~\cite{Mlakic2018}, the authors use a CNN to analyze thermal images by looking at the strongest activation channel of the intermediate layers to construct a visual representation of the early signs of failure in power transformers.
In another impressive application, Lakhani et al.~\cite{Ziletti2018} are able to visualize features on the intermediate layers in a CNN to detect pulmonary tuberculosis on chest radiographs. An additional example illustrates the use of the activations of the early layers to  highlight driver behaviours~\cite{Xing2019}.

The singular value decomposition (SVD) of a matrix $A$ representing a collection of images, provides with an orthonormal basis that can be used to describe correlations between individual images in $A$. 
We use SVD to discover the features in X-ray images that are relevant for classification. 
Although SVD has been used to generate the input to a neural network~\cite{Lawrence1997,RamaLingaReddy2010} and to  improve the training of the networks~\cite{Zhang2015,Astrid2018,Wang2020}, to the best of our knowledge, the analysis of the statistical correlations of the feature maps in the early layers of a CNN has not been utilized to discriminate X-ray images of samples of epoxy resins.

This paper is organized as follows. In Section 2 we describe the experimental data and the properties of the materials.
We provide with a description of the CNN employed in this work and we  highlight the importance of a robust approximation to the gradient field of the X-ray images as a preprocessing step before training the network.
In Section 3 we present two approaches to visualize  characteristic features in the images that are relevant for classification. One method consists in finding the response map with the strongest activation  and the other method leverages the hierarchical ordering of the eigenvectors of a whole set of response maps in a CNN to produce a visual representation of the features that are relevant for classification.
We close with a brief  argument to support the use of a single eigenvector to describe a library of response maps and a discussion of the advantages of the proposed approach.

%METHODS
\section{Methods}	
\subsection{Materials and image acquisition}
\begin{figure}[h]
     \begin{center}
           \includegraphics[width=110mm]{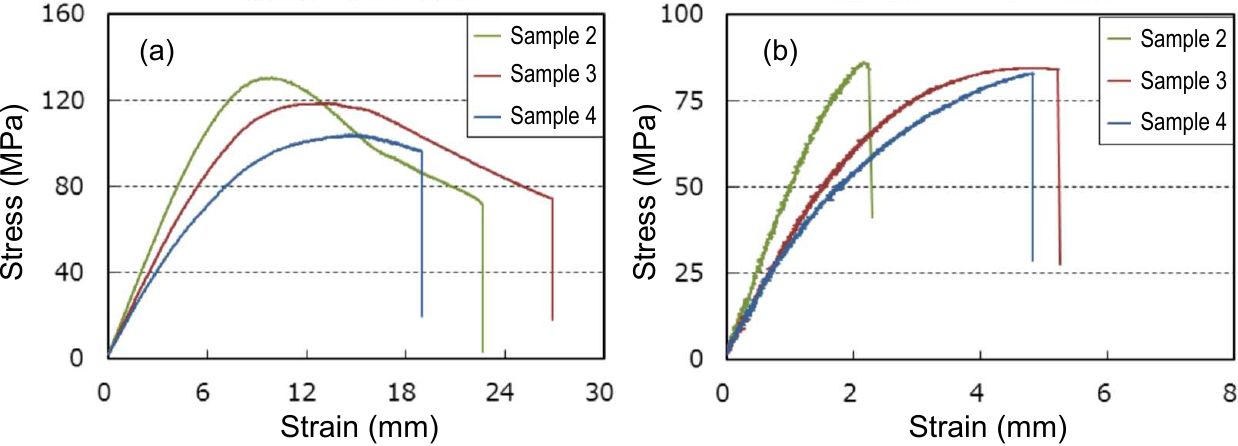}
    \end{center}
    \caption{
    Stress-Strain diagram determined by (a) bending test and (b)  tensile test. In both cases sample No. 3 has the highest total absorbed energy, this is, the area under the s-s curve shown in Table 1. The sample No. 1 is rather fragile and could not withstand the mechanical test.
    Data courtesy of NIPPON STEEL Chemical \& Material CO., LTD.}%
   \label{fig:SS}
\end{figure}

%
% Nishiura request May 11th
%Materials scientist are more interested in the [*performance*],
%We should describe in the first paragraph that the [*difference of process is crucial to its performance*] and we did some analysis in [2]. 
%But we were [*not able to visualize the key feature*] in a clear manner, although we extracted the quantity [*TV as an good indicator*],
%but it was still difficult to see it by our eyes.
%Therefore our [*goal is *]......
%Table 1 is cited here, however it does not indicate that [*sample 3 is the best one*], since it lacks [*strain-stress curve*] for those samples.
%I think we can borrow several expressions and figures from our previous paper [2].
%--end of Nishiura request May 11th
%-----------------------------------------------
In this section we briefly describe the materials used in this study.
%We consider X-ray CT images of four different samples of thermosetting resins as shown in figure~\ref{fig:matImages}.
%
Four different samples of thermosetting resins were prepared and characterized by NIPPON STEEL Chemical \& Material CO., LTD. 
These samples have the same chemical composition of bisphenol A type epoxy molecule and hardener molecule (primary diamine) with a ratio 3:1. 
Different conditions of heating temperature and heating time endow the samples with different polymerization rates, densities, and fracture toughnesses,
as it is shown in Table~\ref{Tab:tableA1}.
The samples were prepared as a plate with a thickness of 1mm and their mechanical properties were measured by the standard means of evaluation~\cite{Swallowe1999}. 
The stress-strain (s-s) curve for the samples in different settings shown in Fig.~\ref{fig:SS}
indicates that the sample No. 3 possesses the best performance in terms of the absorbed energy up to fracture, that is, the largest area under the s-s curve.
The numerical values of the areas are presented in Table~\ref{Tab:tableA1}.
The samples Nos. 2 and 3 possess values of fracture toughness that seem to be close to one another; the experimental error in the measurements is unclear.

\begin{figure}[h]
     \begin{center}
            \includegraphics[width=100mm]{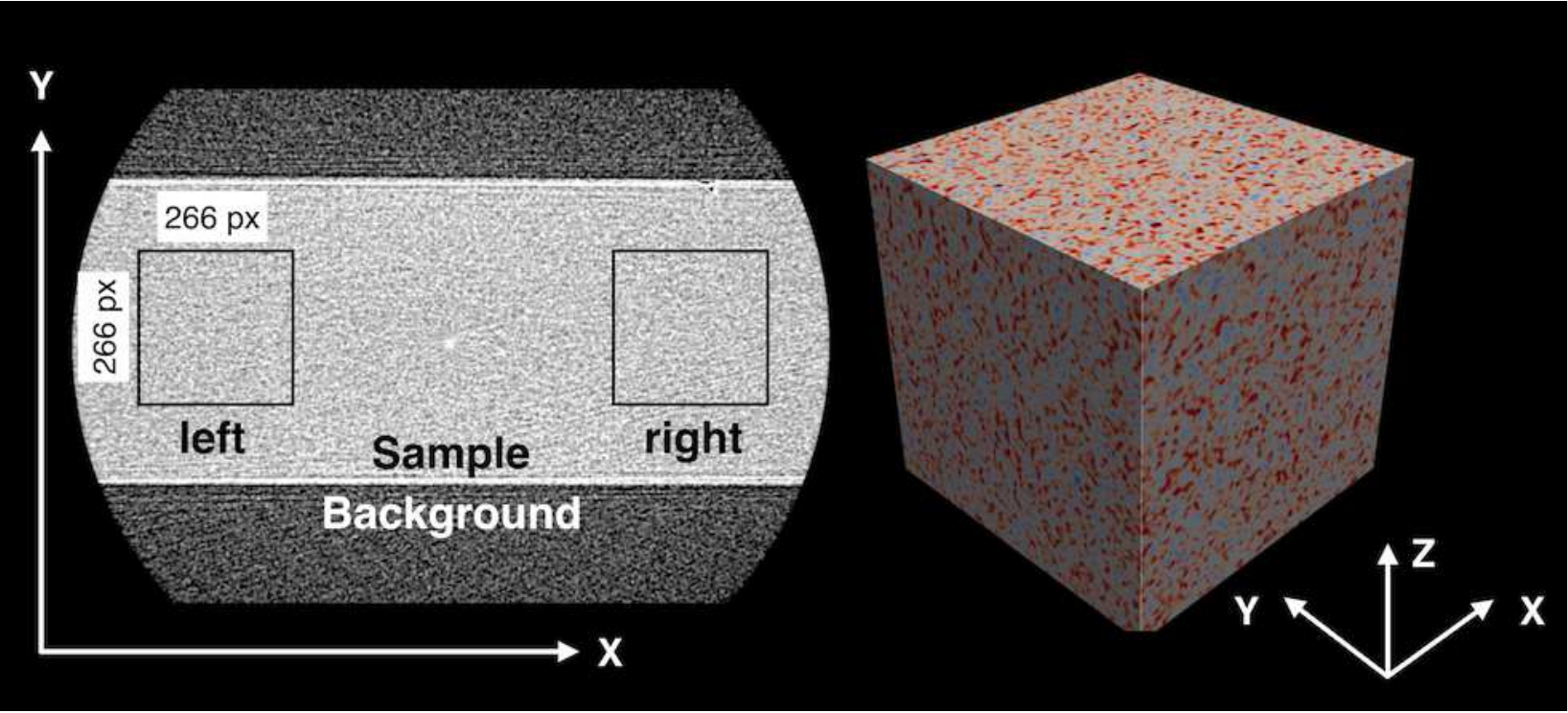}
    \end{center}
    \caption{(Color online)%    
(Left) An example of X-ray CT slice image. The middle bright area corresponds to the sample. The upper and lower dark background area shows a noisy image originated from the fluctuation of X-ray intensity. 
(Right) 3D reconstruction using 266 slices. The brighter (darker) parts are colored red (blue).
     }%
   \label{fig:f03}
\end{figure}

\begin{figure}[h]
     \begin{center}
         \includegraphics[width=73mm]{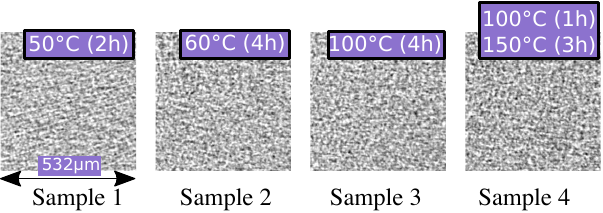}
    \end{center}
    \caption{%
X-ray CT images of four samples of epoxy resins. 
Variations of the process variables, such as heating and heating time, produce different patterns of density in the X-ray images. 
 }%
   \label{fig:matImages}
\end{figure}

In addition to the mechanical tests, 
 X-ray CT images of all samples were obtained using a commercial apparatus with the resolution of 2 $\upmu$m$/$pixel.
 % by NIPPON STEEL Chemical \& Material CO., LTD.
 %
 Two squared sections of size $266\times266$ pixels  were cut out from left and right sides of the plate, which were then sampled into X-ray images every 2 $\upmu$m (Fig.~\ref{fig:f03}).  
A total of 266 slices of each sample were extracted and thus our dataset consists of 1064 X-ray images.
Some representative X-ray CT  images of samples 1-4 are shown in Fig.~\ref{fig:matImages}.
In these images, the bright areas correspond to regions of high electron density ($ \fallingdotseq$ high atomic density).
Since the three-dimensional network of covalent bonds obstructs the packing of molecules, 
we assume that the highly polymerized regions correspond to low density (dark) domains.
On the other hand, the highly polymerized region has a larger Young's modulus, therefore, 
the dark region is considered to have a larger Young's modulus than the bright region. 
%
% new ***
From this observation it follows that variations of the patterns of the intensity of the original images represent variations in the elastic modulus.
Based on numerical simulations and theoretical considerations using the phase field framework described in Ref.~\cite{Avalos2020},
 it was shown  that high values of the TV of the elastic modulus (or equivalently  high values of the total gradient content of the images) represent  high values of effective toughness.
%*********
%As it was mentioned above, variations of  the patterns of the intensity of the X-ray images  have a significant impact on the   mechanical properties of the samples, which is revealed by
%the finding in Ref.~\cite{Avalos2020}, that shows the TV (or equivalently the total gradient content of the images) is an appropriate parameter to describe the samples in terms of their mechanical performance.
%
However, variations in the effective toughness  are not recognizable by simple observation. Therefore our goal is to identify the visual fingerprints in the samples that are relevant for their classification.
%-----------------------------------------------

%
\begin{table}[ht]
\centering
\caption{Properties of samples 1-4. Data courtesy of NIPPON STEEL Chemical \& Material CO., LTD.
Areas under the s-s curve normalized with the area of their respective domain according to Fig.~\ref{fig:SS}.}
\begin{tabular}[t]{lcccc}
\toprule
~ & Sample 1 & Sample 2 & Sample 3 & Sample 4\\
\midrule
Polymerization Rate  {[\%]} &13.7& 45.9& 60.3& 90.0\\
Density  {[$g/cm^3$]}  &1.151&1.157& 1.145&1.143\\
Fracture Toughness {[}MPa$\cdot$m$^{1/2}${]}&0.16&1.03&0.99&0.83\\
Area (tensile test) {[a.u.]}  & -- & 0.91 &  2.46 & 2.02\\
Area (bending test) {[a.u.]}  & -- & 0.44 &  0.51 & 0.31\\
\bottomrule
\end{tabular}
\label{Tab:tableA1}
\end{table}%

%

\begin{figure}[h]
     \begin{center}
            \includegraphics[width=73mm]{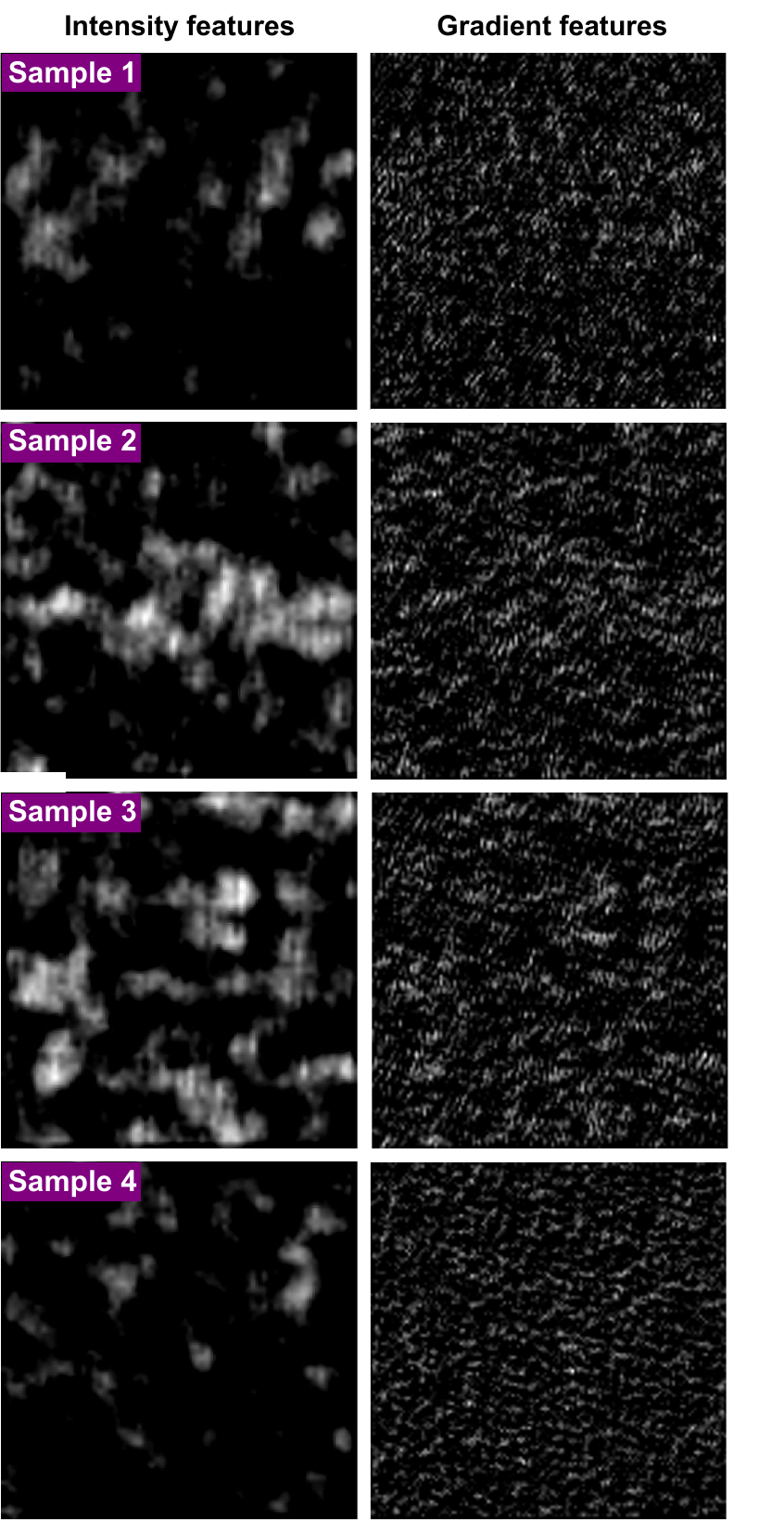}
    \end{center}
    \caption{%    
Feature maps of samples  1-4 (top to bottom) obtained  by training the CNN described in section~\ref{cnn} and then the features of a test image of samples 1-4 are extracted.
The first column shows results  obtained directly using the original X-ray images of the intensity field, $Y$.
Notice that the features lack of well-defined boundaries and occupy large areas of the domain.
The preprocessing presented  in section \ref{gradient} considerably improve  the visualization of the features rendering more accurate responses. 
The second column shows results obtained from transformed X-ray images, $\nabla Y$, using the strongest channel, $\alpha$, as presented in Eq.~\eqref{eqn:strongestActiv} in section~\ref{featuremaps}.
The features vary considerably when the initial conditions of the training are changed. 
In  section \ref{deepeigenfeatures} we develop a more robust method to extract features.
Compare these feature maps with the richer and more accurate details shown in Fig.~\ref{fig:inverseProbEigen}.
}%
   \label{fig:robustFailure}
\end{figure}

%%%%%%%%%%%%%%%%%%%%%%
%\textcolor{red}{
In section~\ref{cnn} we use a neural network to classify images like those  shown in Fig.~\ref{fig:matImages}.
Our goal is to discover what are the most distinctive  features in the X-ray images that are used to classify these images into four classes.
However, the direct use of the original  X-ray images is delicate because the images contain a noisy background resulting from the  measurement process. Although it is possible to  use  a neural network to extract the features directly from the original images of the intensity field,  the results are not satisfactory, as it is shown in the first column of Fig.~\ref{fig:robustFailure}, which exhibits large areas of activation with unclear indication of the features.
%The deficient definition of visual details is the upshot of noise in the images.
In order to render a more faithful description of the features, it is desirable to preprocess the X-ray images to highlight more clearly the variations of the intensity field.
%}

%\textcolor{red}{
A first task is to determine what variable is adequate to produce a good visualization of the features of the images.
It was previously found~\cite{Avalos2020} that, for a given image with intensity field $Y$,   the magnitude of the gradient, $|\nabla Y|$,  seems adequate to
preprocess the X-ray images because the average value of the gradient field, $\overline{|\nabla Y|}$, describes the samples 1-4 according to their mechanical performance.
For example,  Fig.~{\ref{fig:stats}b} shows that
the quantity  $\overline{|\nabla Y|}$ is larger in  the sample No. 3, which also possesses  the largest area under the s-s curve, as it is shown in Fig.~\ref{fig:SS} and in Table 1.
On the other hand, although $\overline{|\nabla Y|}$  allows to describe the samples according to the mechanical performance, it would be desirable to develop a method to  visualize the most important features on the images for classification. The present work is an attempt to develop such method of visualization.
%Thus, $\overline{|\nabla Y|}$ describes the samples 1-4 according to their mechanical performance,

\begin{figure}[h]
     \begin{center}
            \includegraphics[width=65mm]{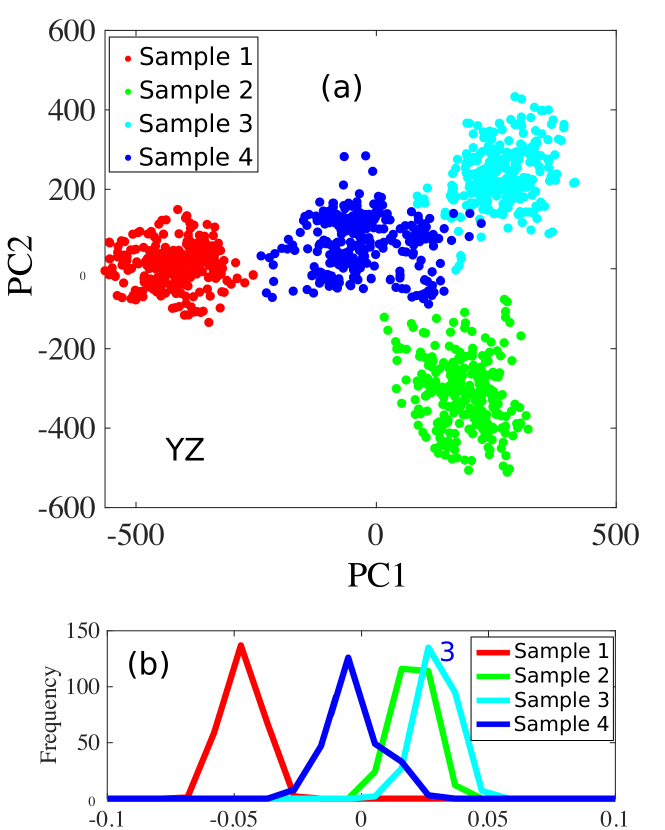}
    \end{center}
    \caption{%
    Adapted from Ref.~\cite{Avalos2020}.
    (a) Projection of the gradient of the original images in the plane YZ onto PC${}_1$ and PC${}_2$ for slices.
    (b) Histograms of the distribution of loadings for the four samples on the first dominant SVD mode.
 %The loadings come from the column of the $V$ matrix of the SVD.
 %
 The ordering of the clusters along the first dominant SVD mode suggests that sample No. 3  contains the largest value of the total variation. Notice that cluster of sample No. 3  appears on the rightmost location followed by samples with labels 2, 4 and 1. 
     }%
   \label{fig:histogram}
\end{figure}

The  gradient field is adequate  to  preprocess the X-ray images for several reasons.
Firstly,
the gradient is used frequently in industrial inspection, either to aid humans in the detection of defects or, what is more common, as a preprocessing step in automated inspection~\cite{Essid2018}. Additionally the ability to enhance small discontinuities in an otherwise flat gray field is another important feature of the gradient~\cite{Gonzalez2006}.
%the features seen in a digital image have boundaries that can be obtained by detecting changes in intensity~\cite{Gonzalez2006}.
%The magnitude of the gradient of an image highlights  the boundaries of the features.
Secondly,
although a convolutional neural network can use the original X-ray images to extract characteristic features of the images, these appear blurry and not well defined, as  it is shown on the first column of Fig.~\ref{fig:robustFailure}. An intuitive explanation is that while some convolutions taking place at the first layers of the neural network are approximately similar to the gradient calculation, the  individual loadings responsible for the convolution are not specifically designed to compute the gradient.
The loadings at the layers are meant to distinguish different kinds of features on the images %, not only variations of intensity.
and are optimized during the network training, but they are not intended to function as a robust discretization of the gradient operator.
In contrast, a good gradient operator is designed to produce a robust description of the variations of the intensity field. In the next section we employ a gradient operator that is robust under noisy conditions.
Thirdly,
the evidence shows that the intensity field is not the appropriate variable to describe the mechanical performance.
Previously it has been show that  the intensity does not provide the correct sequence of the projections of the X-ray images onto a eigenspace spanned by the first two principal components.
However the projections of  the  gradient images, $|\nabla Y|$,  have  the correct sequence, in which sample No. 3 has the best mechanical performance, as it can be seen in Fig.~\ref{fig:histogram}. 
This figure shows that sample No. 3  appears on the rightmost location followed by samples 2, 4 and 1.
Full details of how this clustering was obtained appear in Ref.~\cite{Avalos2020}. 
For the purpose of this work, we  recall that projecting the magnitude of the gradient of the original images onto a space spanned by principal components, preserves the ordering of the samples according to their ability to absorb energy up to fracture. In other words,
computing the gradient images allows us to correctly  describe the toughness of samples 1-4.
%

%********************
%********************
%} 
%

%%%%%%%%%%%%%%%%%%%%

\subsection{Preprocessing} 
\subsubsection{Basic statistical quantities.}
\begin{figure}[h]
     \begin{center}
       \subfigure[Mean of each slice.]{%
            \label{fig:statsA}
            \includegraphics[width=73mm]{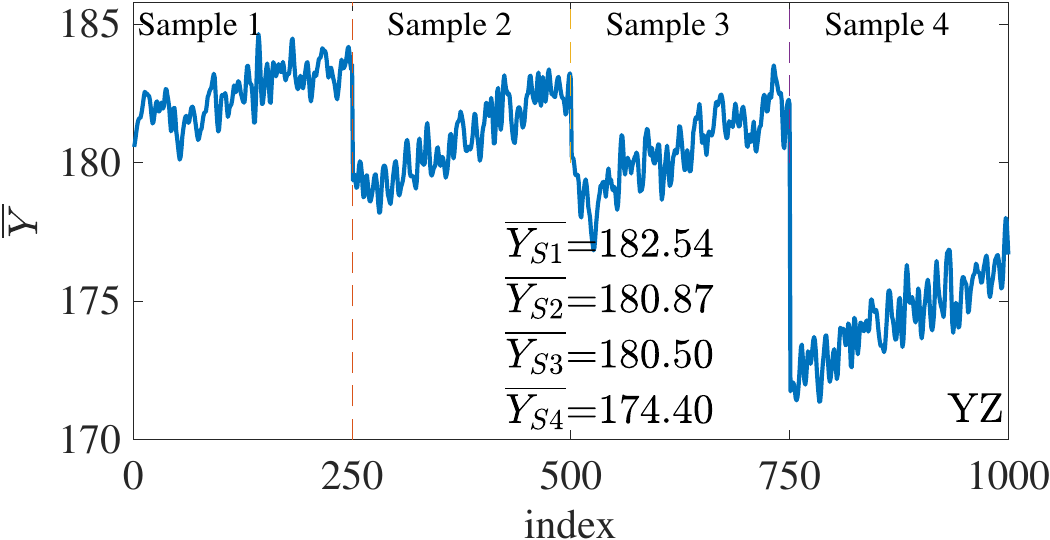}
        } \\  
        \subfigure[Module of the gradient]{%
            \label{fig:statsB}
            \includegraphics[width=73mm]{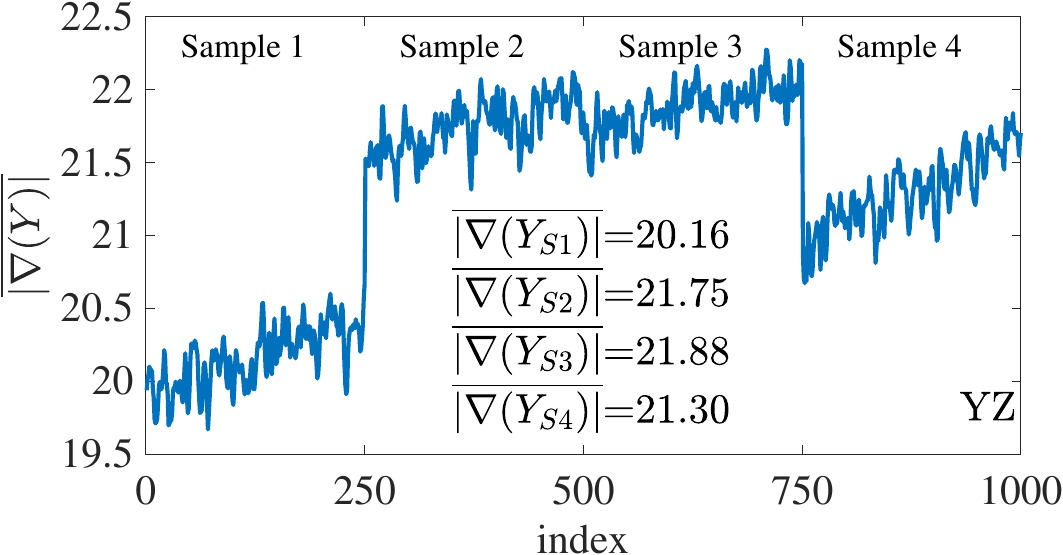}
        } %\\

    \end{center}
    \caption{%
(a) Mean value of the intensity field of each X-ray image  in the plane YZ shown as $ \overline{Y_{Si}}$. 
     The index number represents  each image and runs from 1 to 1000. From left to right,  the vertical lines separate samples 1 to 4.
 (b) Mean of the module of the gradient of each slice, $ \frac{1}{N}\int (f_x^2+f_y^2)^{1/2} dxdy$, with $N$ being the pixel count.
 Notice that sample No. 3  has the largest average value of the module of the gradient followed by samples with labels 2, 4 and 1.
     }%
   \label{fig:stats}
\end{figure}

We look into some basic statistical parameters in the set of the X-ray images. 
To analyze the distribution of the pixel values in the X-ray images, we proceed as follows. For each grayscale X-ray image with pixels taking values between 0 and 255,  we compute the mean value in a fixed squared domain ($266\times 266$). The total intensity of the images divided by the pixel count corresponds to the mean value of the images.
Figure~{\ref{fig:stats}a} shows the mean value of  the slices of  samples 1 to 4, and %
Fig.~{\ref{fig:stats}b} shows the mean of the module of the gradient of each slice computed as $ \frac{1}{N}\int (f_x^2+f_y^2)^{1/2} dxdy$, with $N$ being the pixel count. Notice that sample No. 3  has the largest average value of the module of the gradient followed by samples 2, 4 and 1.

\clearpage
\newpage

\subsubsection{Robust gradient approximation} \label{gradient}
The original  X-ray images  describing  the intensity field  contain large fluctuations of intensity and it is required to remove all unnecessary information from the images.
Transforming the intensity field into the module of the gradient  highlights the features of the X-ray images. 
More importantly,  the summation of all the local  contributions of the module of the gradient --the total variation-- is a quantity related to the mechanical performance  of the samples~\cite{Avalos2020}.
The role of this transformation is additionally strengthened by the fact that the module of the  gradient produces the correct classification of the images,
in which the sample No. 3  has the largest average value of the module of the gradient followed by samples with labels 2, 4 and 1~\cite{Avalos2020}.

To  compute the module of the gradient we process as follows.  For each image $Y$ defined as a $m\times n$ array, we need to compute the derivatives in the vertical  and horizontal directions, $\frac{\partial Y}{\partial x}$ and $\frac{\partial Y}{\partial y}$, respectively.  The module of the gradient is then defined as
\begin{equation}
{\nabla Y}=
\left[
  \begin{array}{c}
    Y_x \\
    Y_y  
  \end{array}
\right]
=
\left[
  \begin{array}{c}
    \frac{\partial Y}{\partial x} \\
    \frac{\partial Y}{\partial y} 
  \end{array}
\right]
\end{equation}
\begin{equation}
|{\nabla Y}|=\sqrt{ Y_x^2+Y_y^2},
\label{eqn:nablaY}
\end{equation}
where $Y_x$ and $Y_y$ represent simple discrete approximations of the forward difference for interior data points in the the vertical  and horizontal direction, respectively. For example, for a matrix with unit-spaced data, $Y$, that has vertical gradient, $Y_x$, the  interior gradient values are computed as $(f_{i,j+1}-f_{i,j})$.  The  horizontal gradient is computed similarly.

\begin{figure}[h]
\centering
    \subfigure[ $|{\nabla Y}|$  computed using forward difference.]{
    \label{fig:uno}
\includegraphics[width=55 mm]{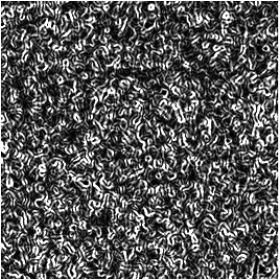}
}
    \subfigure[ $|{\nabla Y}|$  computed using a Sobel mask.]{
    \label{fig:dos}
\includegraphics[width=55 mm]{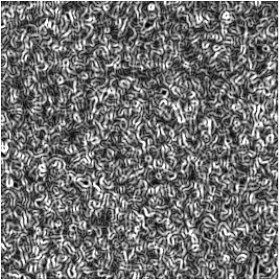}
}
\caption{
Image of module of gradient of a typical sample using two different discretization methods.
The intricate texture of the images is the result of the local variations of the intensity field captured by the transformation.
}
    \label{fig:twomasks}
\end{figure}

The calculation of the module of the gradient of each image $Y$ shows that the transformed image $|{\nabla Y}|$ reveals previously unseen intricate variations of the intensity field  as it is shown in Fig.~\ref{fig:twomasks}.
%(ii) The ordering  of the clusters of new images in the absolute gradient space, places the sample 3 as the best performing material.
Additionally, it is well-know that noise suppression is an important issue when dealing with derivatives to compute the gradient~\cite{Gonzalez2006}. 
Therefore the discretization of the gradient function requires special attention. 
Although simple algorithms of differentiation such as central difference and forward difference  produce good results of clustering~\cite{Avalos2020}, these methods  are  not longer satisfactory to visualize the gradient of an image with noisy background. 
A more comprehensive  computation of the gradient  is necessary to capture a richer content of visual details  in the image of $|{\nabla Y}|$.

\begin{figure}[h]
\begin{center}
\includegraphics[width=90mm]{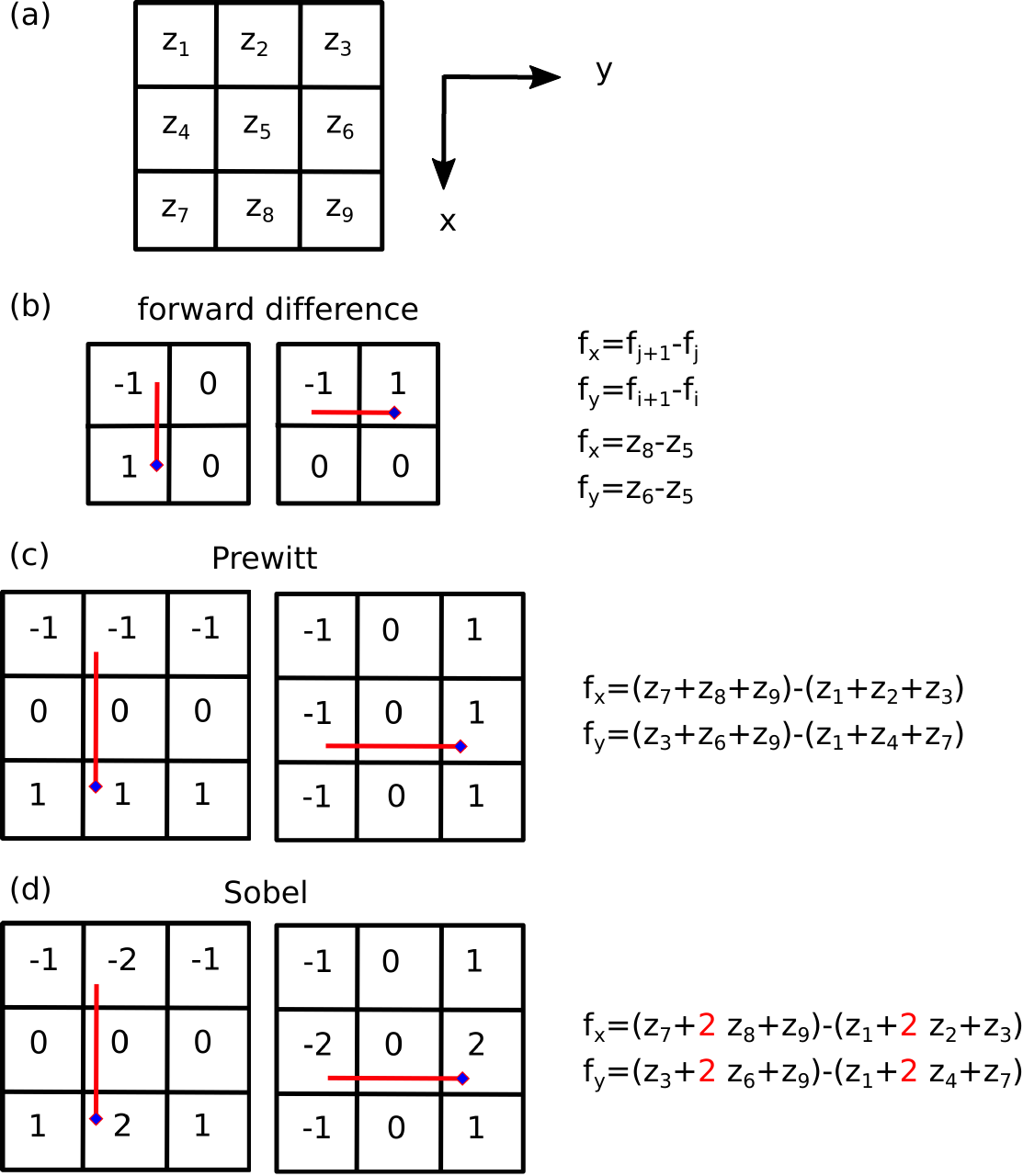}    
\end{center}
\caption{
Filter masks to compute the derivatives needed for the gradient operator.
(a) Generic $3 \times 3$ mask  with the ordering of its elements shown as $z_p$, with $p=1,..,9$.
(b) Simple forward approximation,
(c) Prewitt mask and
(d) Sobel mask. Notice this mask gives twice the weight to the central pixel compared to Prewitt.
}
\label{fig:masks}
\end{figure}

In addition to  forward differences, there are other methods to approximate the gradient.
More in general, the  gradient of a given image $Y$ is computed through a 2D convolution with a $3\times 3$ mask $Z$, as shown in Eqn.~\ref{eqn:2dConv}.
\begin{equation}
C(x,y)= \sum_{t=-1}^1\sum_{s=-1}^1   Z(s,t) Y(x-s, y-t)
\label{eqn:2dConv}
\end{equation}
Figure~\ref{fig:masks} shows several masks to perform the approximation of the derivatives needed for the gradient operator.
While the forward  difference approximation shown in Fig.~\ref{fig:masks}b preserves clusters in an eigenspace~\cite{Avalos2020}, the  contribution of additional  neighbouring locations improves the gradient approximation.
The Prewitt mask shown in Fig.~\ref{fig:masks}c, produces a more comprehensive  value of the gradient that includes neighbouring  contributions of a given location.
Additionally, the Sobel mask  shown in Fig.~\ref{fig:masks}d gives more weight to the central pixel and also has better noise-suppresion (smoothing) characteristics~\cite{Gonzalez2006}.
For a given image $Y$, the derivatives $Y_x$ and $Y_y$ can be computed using Eqn.~\ref{eqn:2dConv}  with the  Sobel masks shown in Fig.~\ref{fig:masks}d, and
then $|{\nabla Y}|$ can be found from  Eqn.~\ref{eqn:nablaY}, which is the absolute gradient  field of the image $Y$.

As a visual example, Fig.~\ref{fig:twomasks} shows $|{\nabla Y}|$ for a typical image of sample 1 computed using  forward difference and a Sobel mask. Notice that additional details are captured by the Sobel mask.
In what follows, we use the Sobel operator to transform the intensity field of the whole set of the X-ray images into the corresponding absolute gradient field. The goal is to discover the most notable features that are relevant to classify the images.
% For this figure I used sample 1
% %%%%%%%%%%%%%%%%%%%%%%%%%%%

%\clearpage
%\newpage

\subsection{Convolutional neural network} \label{cnn}

\subsubsection{Architecture description }\label{archCNN}
A CNN is a type of neural network~\cite{Fogelman-Soulie1987,Haykin2009} that uses convolution (Eqn.~\ref{eqn:2dConvGeneral}) to process 2D  numerical arrays such as images.
%
 %
% Figure~\ref{fig:cnnScheme} shows  a schematic representation of the CNN used in this work.
 We employ a CNN composed of several convolutional blocks including convolution, batch normalization, rectified linear unit (ReLU) and maxpooling. 
 The final end of the  CNN consists of one  fully-connected layer and one Softmax layer.
 The individual components of the CNN are described next.

 The two-dimensional convolution of the input $Y$ with a kernel $W$, denoted as $C(m,n)=Y(m,n)*W(m,n)$, is defined as
 \begin{equation}
C\left(m,n \right)= \sum_{t=\infty}^{\infty} \sum_{s=-\infty}^{\infty}   W\left(s,t \right) Y\left(x-s, y-t \right)
\label{eqn:2dConvGeneral}
\end{equation}
Where $m$ and n are indices along the horizontal and vertical directions, respectively.
The resulting function $C(m,n)$ is the output or feature map of the convolution.
For a given image $y_i$ at the $l$-th layer, the convolution produces the output
 \begin{equation}
s^l_j=f\left(y^{l-1}_i * w^l_{ij}+b^l_j\right)
\label{eqn:convOutput}
\end{equation}
Where $w^l_{ij}$ are learnable parameters, $b^l_j$ is a bias and $f(\cdot)$ is the output of the activation function~\cite{Bouvrie2006}.
We implement batch normalization~\cite{Ioffe2015,Goodfellow2016,Santurkar2018}   right after each convolution and before activation by calculating the mean and standard deviation of each input variable to a layer per mini-batch, 
which consists of a subset of the training dataset.
%
%We adopt batch normalization~\cite{Ioffe2015,Santurkar2018}  right after each convolution and before activation to standardize each input variable to a layer per mini-batch that consists of a subset of the training dataset
%
%To normalize the activations of the convolution, $x_i$, we use batch normalization~\cite{Ioffe2015,Santurkar2018}
We employ a mini-batch mean $\mu_B$ and standard deviation $\sigma^2_B$   as:
 \begin{equation}
\hat{x_i}= \frac{x_i- \mu_B}{\sqrt{\sigma^2_B+\epsilon}}
%\hat{x_i}= \frac{x_i- \mu_B}{\sqrt{\sigma^2_B+\epsilon}}
\label{eqn:batchNorm}
\end{equation}
 A small value of $\epsilon$ is needed in case the mini-bach variance is very small.
The rectified linear unit (ReLU) defined in Eqn.~\ref{eqn:relu}~\cite{Jarrett2009,Nair2010}, is employed as nonlinear activation function because it is faster than other alternatives such as the sigmoid or $\tanh(\cdot)$ functions. What the ReLU function does is to clip the negative part out of the output of the activations of the convolution.
 \begin{equation}
g(z)=max \left(0,z \right)
\label{eqn:relu}
\end{equation}
Maxpooling is employed on the activations after the  ReLU  function to  down-sample the spatial size of the input arrays by taking the maximum values of a subset of the input array~\cite{Nagi2011}.

\begin{figure}[h]
\begin{center}
\includegraphics[width=110mm]{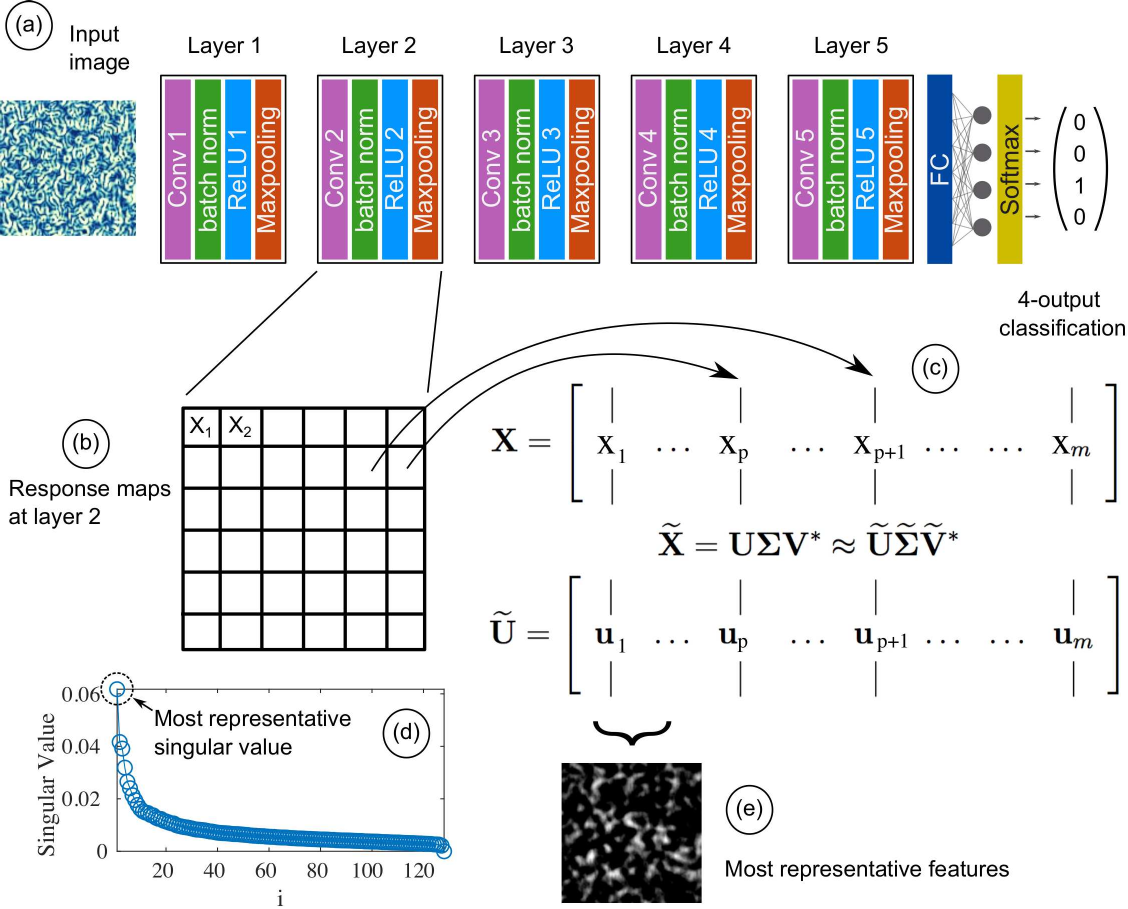}    
\end{center}
\caption{
Schematics of the CNN used in this work with individual components decribed in section \ref{archCNN}.
(a) The gradient of a test image $|{\nabla Y}|$ is the input of a trained CNN. Each convolutional block  consists of convolution, max-pooling, and ReLU. A fully-connected  layer with four outputs at the end.
(b) The second convolutional layer contains 128 channels of feature maps.
  Each feature map ${X}_{i}$  arranged as a column vector goes into  a matrix $\mathbf{X}$.
(c) SVD of $\mathbf{X}$ produces a left singular eigenvector $U$ whose first column $u_1$ contains the most representative features of $\mathbf{X}$.
(d) The fast decay of the singular values indicates that $\mathbf{X}$  has its larger correlation along the $u_1$ eigendirection.
Singular values are obtained from the singular value decomposition of  the matrix $\mathbf{X}$.
(e) The first eigenvector $u_1$ reshaped into an image with size equal to a feature map, ${X}_{i}$, represents the eigenfeatures of the input test image.
The SVD-based method is described in section \ref{secSVD}.
}
\label{fig:cnnScheme}
\end{figure}

Five convolutional blocks are employed in sequence and then the fully connected layer executes a linear combination of the  features  of the output of the previous layer as:
 \begin{equation}
h^l_j=f\left(\sum_i \left(x^{l-1}_i  w^l_{ij} \right)+b^l_j \right)
\label{eqn:fc}
\end{equation}
In this case $l$ denotes the $l$-th layer, $b_j$ is a given bias, $w_{ij}$ is the $ij$-th weight between the input $x_i$ and the $j$-th output unit $h_j$. In our case $j=$1,2,3, and 4.
To normalize the output of the  fully connected layer in the range $[0,1]$, we use the softmax function (Eqn.~\ref{eqn:softmax})~\cite{Bishop2006}.
The values of $\text{softmax}(z)_i$ represent probabilities of the classes $i=$1,2,3, and 4. The total sum of the probabilities of the four classes is 1.
 \begin{equation}
\text{softmax}(z)_i=\frac{e^{z_i}}{\sum_j e^{z_j}}
\label{eqn:softmax}
\end{equation}
% %
We initialize the training with random values of the weights and then run the entire data set of images for several epochs. 
The classification error for multiple samples and multiple classes is computed using cross-entropy~\cite{Bishop1995} as loss function, $J$, as
 \begin{equation}
J=-\sum^N_{i=1} \sum^K_{j=1} t_{ij} \ln y_{ij}
\label{eqn:costFun}
\end{equation}
Where $t_{ji}$ indicates the $i$-th image belongs to the $j$-th class and $y_{ij}$ is the $i$-th prediction obtained for class $j$ from the softmax activation (Eqn.~\ref{eqn:softmax}).
We want to minimize the cost function with respect to all the parameters  $w^l_{ij}$ in the model as 
\begin{equation}
\underset{ w^l_{ij}}{\arg\min} ~~J
    \label{eqn:minJ}
\end{equation}
The details of the architecture  of the  convolutional neural network used in this work are presented in Table 2 and 
a schematic representation of the CNN is shown in Fig.~\ref{fig:cnnScheme}.

%%Please replace the old table 2 with the new one below
%\begin{table}[h]
%\begin{center}
%  \caption{}
%  \label{tbl:cnn-table}
%  \includegraphics[width=70mm]{./figures/table.pdf}
%  \end{center}
%\end{table}
%
% new table 2 (editable)
\begin{table}[ht]
\centering
\caption{Architecture of the convolutional neural network}
\begin{tabular}[t]{ll}
\toprule
Type of layer & Specification\\
\midrule
Convolutional layer & Kernel: 3 $\times$ 3; 128 filters  \\
ReLU & ~\\

Max pooling layer & Pool size: 2 $\times$ 2; stride: 2 $\times$ 2 \\
Convolutional layer & Kernel: 5 $\times$ 5; 128 filters \\
ReLU &~\\

Max pooling layer  &Pool size: 2 $\times$ 2; stride: 2 $\times$ 2 \\
Convolutional layer & Kernel: 7 $\times$ 7; 64 filters \\
ReLU &~\\

Max pooling layer & Pool size: 2 $\times$ 2; stride: 2 $\times$ 2 \\
Convolutional layer & Kernel: 9 $\times$ 9; 64 filters) \\
ReLU &~\\

Max pooling layer & Pool size: 2 $\times$ 2; stride: 2 $\times$ 2 \\
Convolutional layer &  Kernel: 11 $\times$11; 64 filters) \\
ReLU &~\\

Max pooling layer  & Pool size: 2 $\times$ 2; stride: 2 $\times$ 2 \\
Fully connected layer  & size: 4\\
Softmax  &~ \\
\bottomrule
\end{tabular}
\label{Tab:table2}
\end{table}%
%end of table 2
%
%% ejemplo
%\begin{table}[ht]
%\centering
%\caption{Properties of samples 1-4. Data courtesy of NIPPON STEEL Chemical \& Material CO., LTD.
%Areas under the s-s curve normalized with the area of their respective domain according to Fig.~\ref{fig:SS}.}
%\begin{tabular}[t]{lcccc}
%\toprule
%~ & Sample 1 & Sample 2 & Sample 3 & Sample 4\\
%\midrule
%Polymerization Rate  {[\%]} &13.7& 45.9& 60.3& 90.0\\
%Density  {[$g/cm^3$]}  &1.151&1.157& 1.145&1.143\\
%Fracture Toughness {[}MPa$\cdot$m$^{1/2}${]}&0.16&1.03&0.99&0.83\\
%Area (tensile test) {[a.u.]}  & -- & 0.91 &  2.46 & 2.02\\
%Area (bending test) {[a.u.]}  & -- & 0.44 &  0.51 & 0.31\\
%\bottomrule
%\end{tabular}
%\label{Tab:tableA1}
%\end{table}%
%% end

Training was performed using stochastic gradient descent~\cite{Goodfellow2016} to minimize the loss function (Eqn.~\ref{eqn:costFun}) in such a way that the weights of both the convolutional and fully-connected layers are updated to reduce the error.
To evaluate the gradient of the loss function and update the weights, we use batches of 64 images and a learning rate is set equal to $10^{-2}$.
The validation frequency is equal to five iterations.
Using these values,  the classification error  steadily decreases to it smallest value.
 Recent results shows that mini-batches of small size  improved training performance and allow a significantly smaller memory footprint~\cite{Bengio2012,Masters2018}.
 We confirmed that using a mini-batch of size 32 or 64 produces similar feature maps.
 To prevent overfitting, we have employed a dropout layer~\cite{Srivastava2014} at the fully connected layer to randomly set input elements to zero with a given probability of $0.5$.
 Adding a $\ell_2$-regularization term for the weights to the loss function also helps to reduce overfitting~\cite{Bishop2006,Murphy2013}. The loss function with the regularization term takes the form
  \begin{equation}
J_R=J+\lambda \Omega(w)
\label{eqn:L2reg}
\end{equation}
Where $w$ is the weight vector, $\lambda$ is the regularization factor and $\Omega(w)=\frac{1}{2} \| w \|^2=\frac{1}{2} w^T w$. 
For $\lambda > 0$, we minimize $J_R$ as  indicated in Eqn.~\ref{eqn:minJ}.
%we use $\lambda=0.1$
%
Dropout combined with  $\ell_2$-regularization gives a lower classification error~\cite{Srivastava2014}.

%*************
\clearpage

\subsubsection{Image preprocessing and training procedure}\label{ImPreprocess}
\begin{figure}[h]
\begin{center}
\includegraphics[width=100mm]{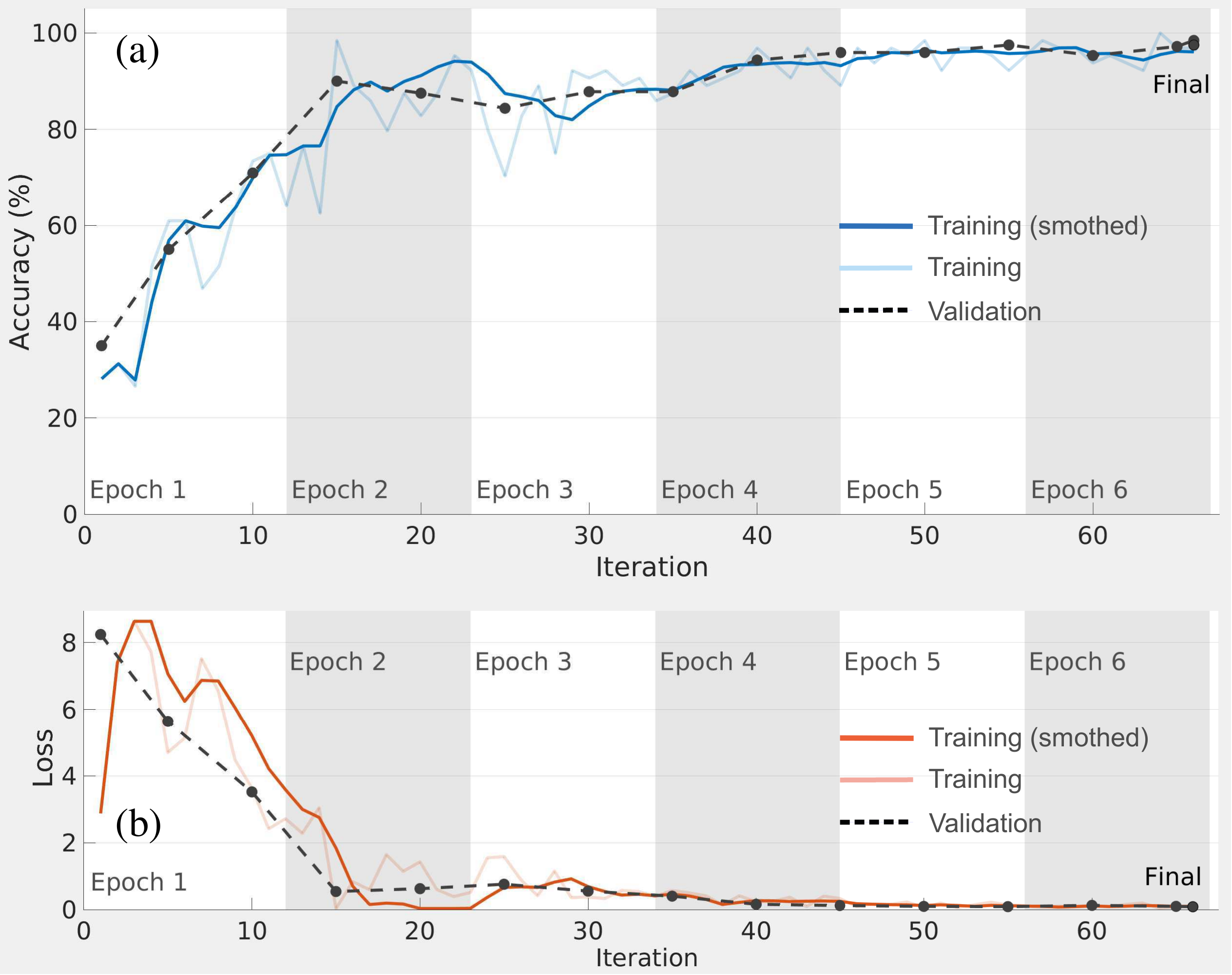}    
\end{center}
\caption{
Training progress of the CNN using the hyperparameters described in the text.
Using a mini-batch size of 64, 
it takes 11 iterations to all of the training samples pass through the learning algorithm.
(a) Training and validation accuracy reaches more or less  the same high values after 60 iterations. 
(b) The steady decrease of the training and validation loss functions suggests that there is not significative overfitting. 
}
\label{fig:training}
\end{figure}

%\textsc{Image preprocessing and training procedure.}
We use grayscale 8-bit images with pixel intensities taking values from 0 to 255.
The set of input images consist of 1064  image files  stored in a TIF format with the resolution of $266 \times 266$ pixels.
To have the data dimensions of approximately the same scale, we  normalize the images  by dividing each image by its standard deviation once it has been zero-centered. Zero-centering means subtracting the mean from each image.
Data augmentation is a powerful method to reduce both the validation and training errors by  artificially in enlarge the training dataset size by  data warping.
The augmented data  represents a more comprehensive set of possible data points, thus minimizing the distance between the training and the validation set~\cite{Shorten2019}.
We employ mirror and upside-down transformations to augment our database. 
The images were grouped in four sets, each with $266$ images and labels of four classes were assigned to all the images.
 For each epoch the training set was randomly divided into 2 groups, 
one data set with 70$\%$ of the images for training and the reminding 30$\%$ for validation. 
The validation data is shuffled before each network validation.
 Over 98$\%$  cross-validation accuracy was achieved.
 Figure~\ref{fig:training}  shows the training progress.
 The network was trained using a CUDA enabled Nvidia Quadro  GP100 GPU.
%

%RESULTS	
  %**************************************************
 \section{Results}
 \subsection{Inverse problem}
  %***************************************************
 The inverse problem consists in finding features in the X-ray images that  are useful to distinguish among  samples 1-4.
 To solve this problem we choose to employ a CNN because these computing systems are able to classify images with high accuracy~\cite{Chollet2018,Jadoon2017}.
 We want to analyze the most representative features that a CNN uses to classify X-ray images.
 Although CNN's  typically achieve a high accuracy in classification,  a deep learning solution to the inverse problem is not easy to interpret because the number of parameters involved in the classification process is of the order of millions.
However a simple approach is to look at the  early layers in the network. 
The filters of the first few layers of a CNN are relatively easy to understand as their primary purpose is  to detect simple details  in the images such as edges.
These edges and other low level features at the  early layers of a CNN describe regions in the images that are important for classification.
In order to achieve a correct classification, the loadings of the filters of  the CNN are optimized during the training process.
%During the training process the loadings of the filters of  a CNN are optimized  for classification
%
%The loadings of the filters of  a CNN are learnable parameters that are optimized during the training process for classification.
%
In this section, firstly we train  a CNN to classify  the transformed images and secondly we extract the learned features at the early layers of the net.
%The schematics of the net used in this work is shown in Fig.~\ref{fig:cnnScheme} and the details of each layer are shown in Table 2.

A  simple idea to address the inverse problem is to feed a gradient image $|{\nabla Y}|$ into a trained CNN and then look at the responses at the early layers~\cite{Chollet2018}. 
The hope is that such responses reveal the presence of  variations of the gradient images across the domain.  These responses at the early layers of a CNN show lumps or spots where the gradient is large and these regions are important for classification.

%***************************************************
 \subsection{Feature maps of a convolutional neural network}\label{featuremaps}
 %***************************************************
 We examine the responses of different layers of the network summarized in Fig.~\ref{fig:cnnScheme} and discover which features the network learns by comparing areas of activation with the original image. 
 Channels in early layers learn simple features such as edges and small hallmarks, while channels in the deeper layers learn complex features. 
 %Identifying features in this way can help understand what the network has learned. 
 We focus only on the  early layers of the network.
 
\begin{figure}[h]
\begin{center}
\includegraphics[width=120mm]{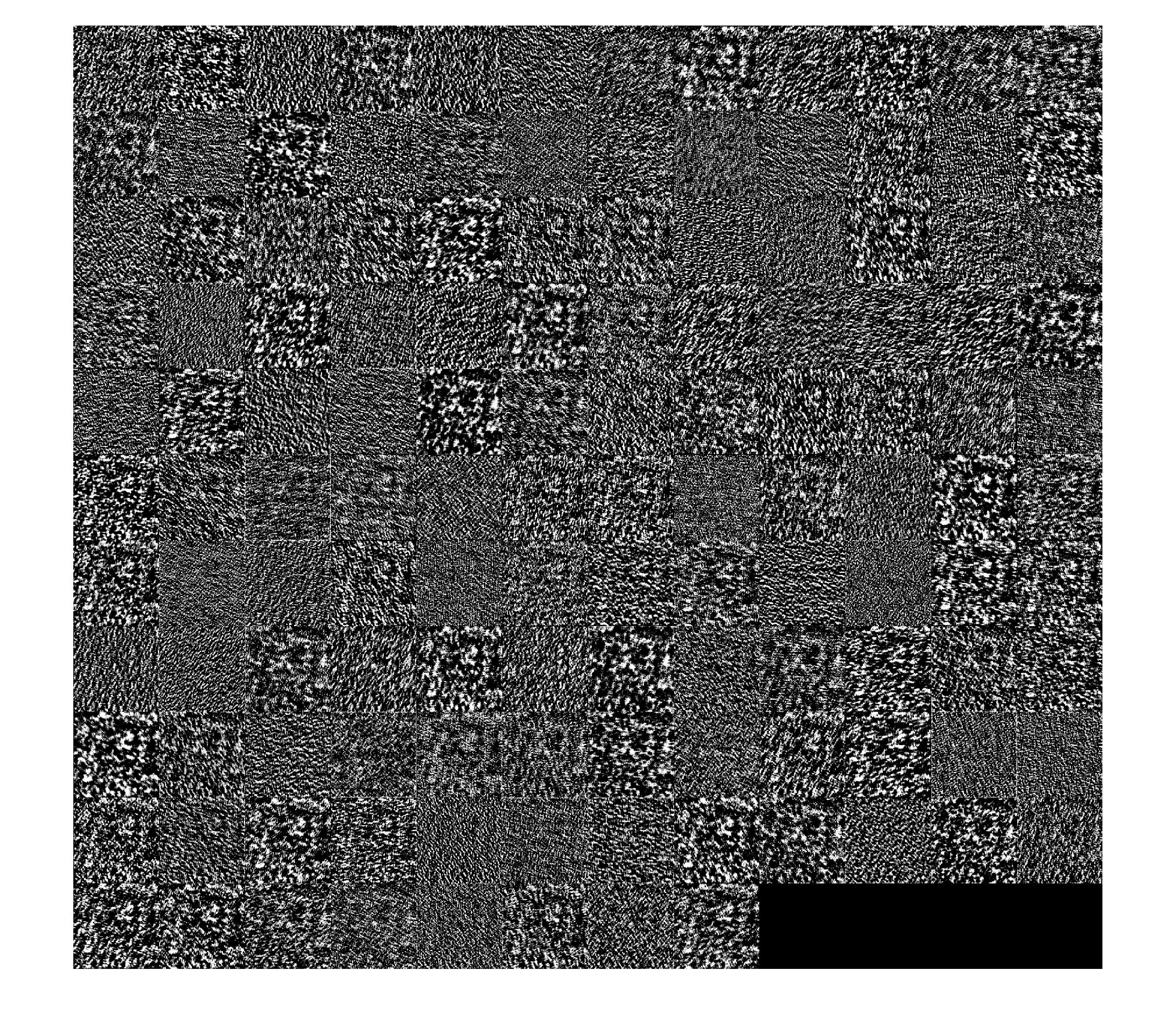}    
\end{center}
\caption{
Feature maps of the 128 channels in the second convolutional layer.
The channels are the response to the classification process of a test image of sample No. 3. The responses at ReLU~2 are shown.
For visualization purposes, the 2$\%$ of all pixel values have been saturated. For the calculations, we employ the original intensity values of each feature map.
The features maps are  arranged in a $12 \times 11$ array. The last four squares at the bottom are empty.
}
\label{fig:featuresMap}
\end{figure}

We consider the first two layers in a CNN, namely, Conv~1 and Conv~2. 
Each layer consists of $3\times 3$ and $5 \times 5$ filters, respectively,
and each of these two layers contain 128 channels of feature maps.
Activation functions  placed after each convolutional layer, are responsible for transforming the summed weighted input from the node into the activation of the node. We use the Rectified Linear Unit (ReLU) as activation function. 
Convolutional  layers Conv~1 and Conv~2 are followed by activation functions ReLU~1 and ReLU~2, respectively. 
 Each channel in the convolutional layers contains a feature map that encodes different responses. For instance, they may encode  diagonal or horizontal edges. Figure~\ref{fig:featuresMap} shows all the features maps at ReLU~2. It is necessary to select a channel that illustrates the learned features at this layer and a common choice  is the channel  with the  strongest activation~\cite{Mlakic2018,Xing2019,Hohman2019}.
 This figure contains 128 features maps arranged in a $12 \times 11$ array. The last four squares are empty.
 To understand the size of one feature map, we recall that the maxpooling operation
 % at the first convolutional block 
 halves the features map from  $266 \times 266$px down to $133 \times 133$px. Then at the second convolutional block the activations at ReLU~2 will have a size equal to $W_2 \times W_2$ with $W_2=(W_1 -F+2 P)/S+1$. 
 Where $W_1=133$ is size of the input, $F=5$ is the spatial extent of the filter, $P=1$ is amount of zero padding and $S=1$ is the stride.
 Then each feature map at ReLU~2 have size $131 \times 131$px and  subsequently is rescaled in the range 0 to 1 and resized to match the  size of the test image, $266 \times 266$ px.
To improve the contrast of  each image in the collage, we have stretched the range of intensity values of each image to span a desired range. We  saturate the upper 1$\%$ and the lower 1$\%$ of all pixel values. This enhancement is only for visualization purposes. For the calculations we employ the original unsaturated image of each feature map.

 To select the channel with the  strongest activation we iteratively search for the strongest activation among all the features maps in the responses at a particular layer.
 % As an example,  we use the second convolutional layer. 
 For two-dimensional feature maps of size $m\times n$ and $k$ channels, the strongest channel is identified with index $\alpha$ is obtained as follows:
\begin{equation}
\alpha=max ({f(x,y,l): x \in 1..m, y \in 1..n, l=1..k})
\label{eqn:strongestActiv}
\end{equation}

\begin{figure}[h]
\begin{center}
\includegraphics[width=66mm]{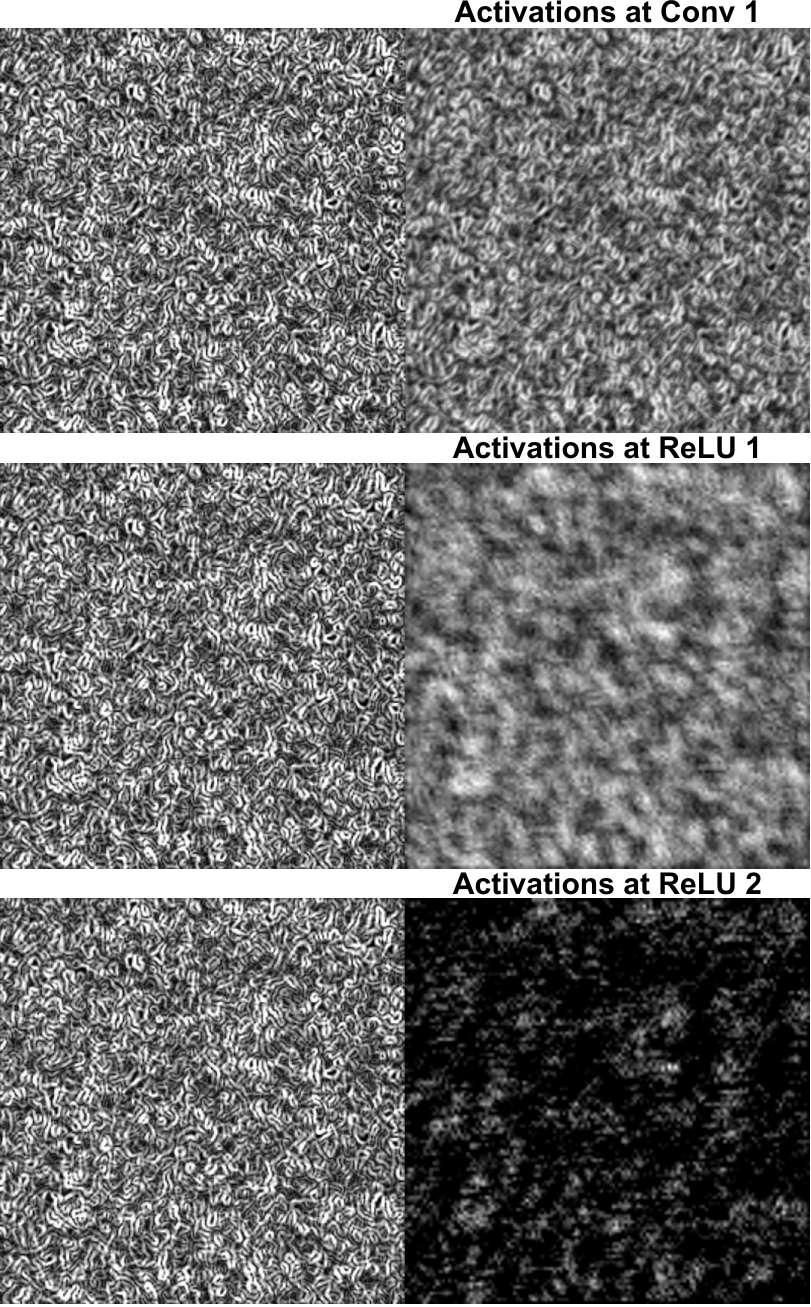}    
\end{center}
\caption{
Comparison of the test gradient image, $X=|{\nabla Y}|$, of sample No. 3 (left hand side) and the channel with the strongest activation (right hand side) at different layers of the network.
From top to bottom, the responses at layers Conv~1, Conv~2 and ReLU~2, respectively.
}
\label{fig:convLayers}
\end{figure}

 Figure~\ref{fig:convLayers}, on the right hand side, shows form top to bottom the  feature map with the strongest activation, $\alpha$, at  Conv~1, Conv~2 and ReLU~2, respectively.
 The left hand side in all cases is the  gradient image, $|{\nabla Y}|$. To simplify the notation we will refer to this quantity as $X=|{\nabla Y}|$.
 Notice that the responses at Conv~1 look quite similar to the original image of  $X$, which means that Conv~1 is fundamentally  acquiring the basic shape of the features of the images.

Conv~2 learns features that are slightly larger  in size than those in Conv~1, because the filters in Conv~2 are slightly larger than those in Conv~1 as well. Notice that Conv~2 reveals  zones where the concentration of  $X$ is large. These zones contains lumps of intensity that can be more easily seen at the activation function of Conv~2, which is called ReLU~2. 
The ReLU function simply clips any negative activation from the output of the second convolutional layer.

\begin{figure}[h]
\begin{center}
       \subfigure[Mean value of  activations at ReLU~1]{%
            \label{fig:meanrelu1}
             \includegraphics[width=73mm]{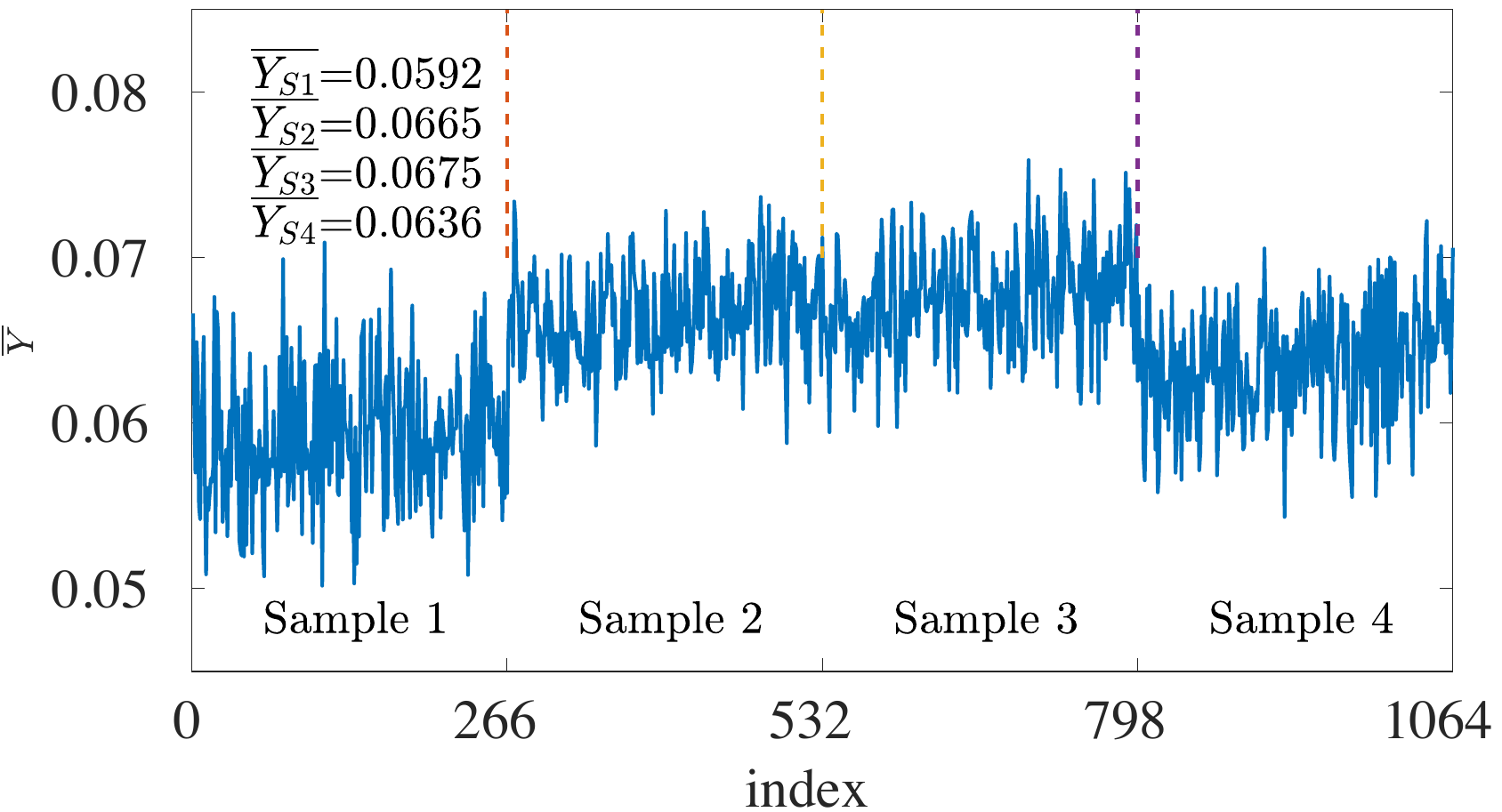}
             }    
        \subfigure[Mean value of  activations at ReLU~2]{%
            \label{fig:meanrelu2}
             \includegraphics[width=73mm]{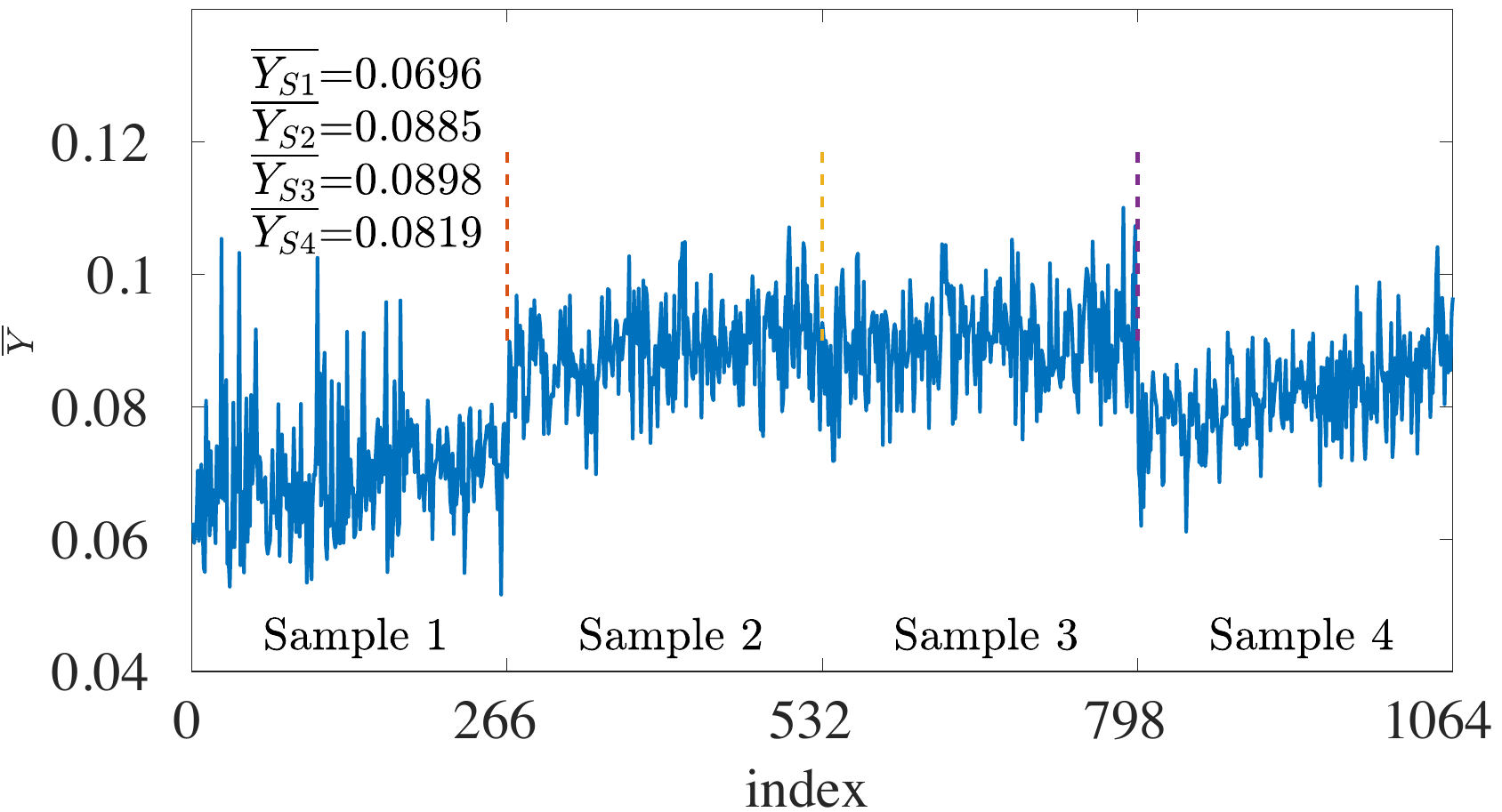}
             }    
             %
%             \subfigure[Mean value of module of the gradient of each slice]{%
%            \label{fig:gradientPaperFig.6d}
%             \includegraphics[width=100mm]{./figures/gradient_t1t2t3t4V2.pdf}
%             }    
\end{center}
\caption{
Average values of the positive activations at the first (a) and second (b) convolutional layers.
%
%(c) Average values of $X=|{\nabla Y}|$ of each slice.   in both cases
%
Notice that  the average values have the same ordering in (a) and (b), and in both cases sample No. 3  has the largest average activation value
followed by samples with labels 2, 4 and 1.}
\label{fig:AverageActivations}
\end{figure}

The bottom row in Fig.~\ref{fig:convLayers} shows a comparison of the original image of $X$ and the $\alpha$ channel  at ReLU~2. The bright lumps highlight zones of large concentration of $X$. 
 One can repeat this process for the whole set of the X-ray images and then compute the mean value of the activations. The result is shown in Fig.~\ref{fig:AverageActivations}. This figure shows the mean value of the $\alpha$ channel  at ReLU~1 and ReLU~2. 
 Notice that  the ordering of these responses agrees with what is presented in Figs.~\ref{fig:histogram} and \ref{fig:stats}b, this is,
 the sample No. 3 appears to have larger activations in average followed by samples with labels 2, 4 and 1 (See also Ref.~\cite{Avalos2020}).
%

%In other words, the mechanical toughness is high in materials with large concentration of the magnitude of the gradient.
%In other words, the activations  describe areas of  with  large concentration of the magnitude of the gradient or high toughness.
 
\begin{figure}[h]
\begin{center}
\includegraphics[width=100mm]{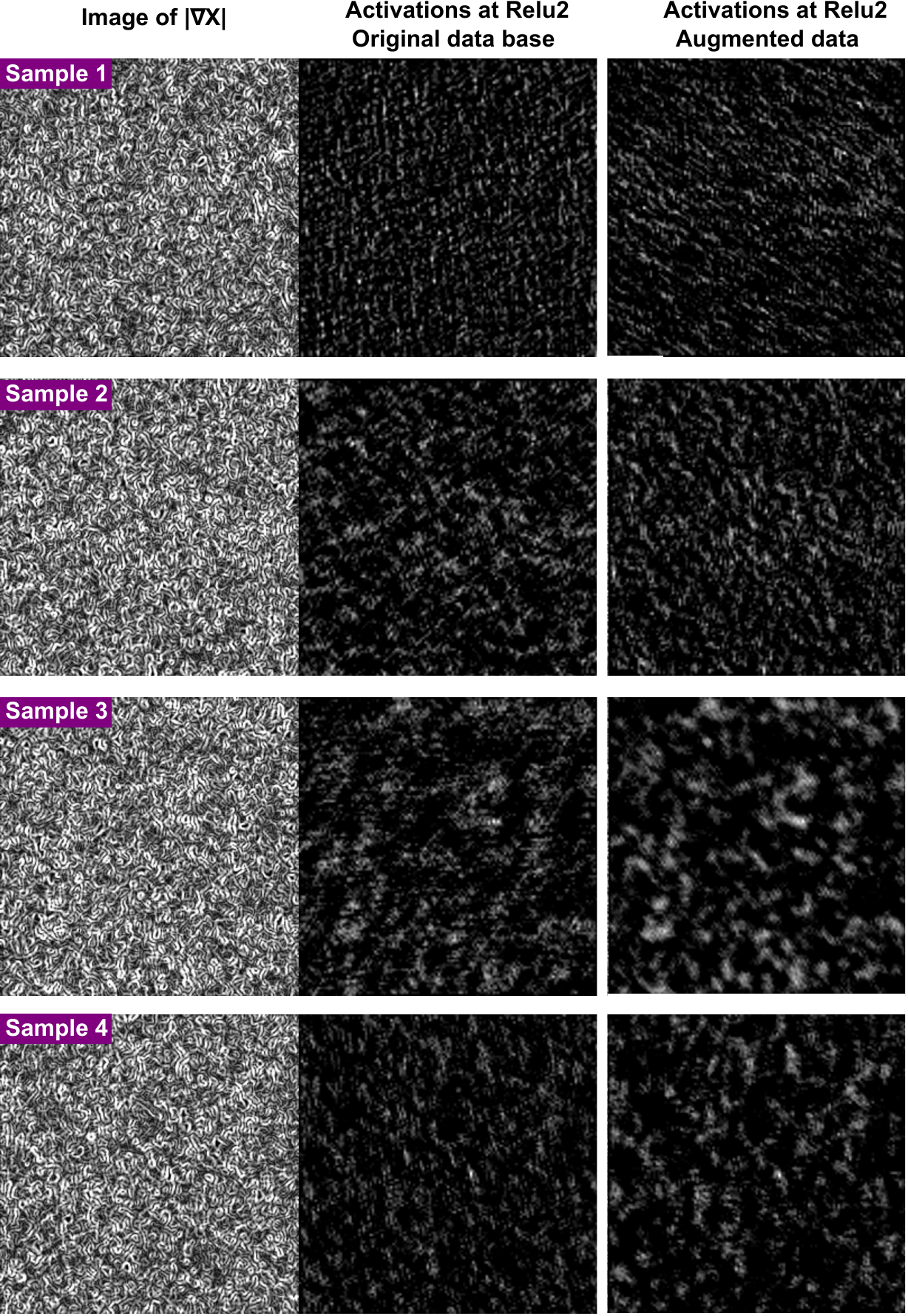} 
\end{center}
\caption{
Features of the test images of samples 1-4 (top to bottom) obtained using the channel with the strongest activation, $\alpha$, at ReLU~2.
The  $\alpha$ channel shows important features in the images for classification. 
First column shows test gradient  images, $X=|{\nabla Y}|$.
Second and third column show features of the images of the first column obtained using  (i) original database and  (ii) the  augmented database, respectively. 
Augmentation is based on mirror and upside-down transformations.
Notice that the responses of sample No. 3 are notably more prominent than those in the other samples, as expected.
}
\label{fig:inverseProb}
\end{figure}

Figure~\ref{fig:inverseProb} is a solution to the inverse problem in which the features in samples 1-4 can be easily distinguished by eye. Notice that sample No. 3  contains a large concentration of  activations. In this figure we employed an augmented database of images, which is constructed by mirror and upside-down reflections on the horizontal and vertical directions, respectively. Using an augmented database artificially  enlarges the training set producing a  clearer distinction  of the  learned features.

%\textcolor{red}{
One of the downsides of using the channel with the strongest activation as  criterion to select the most representative features of each sample is that the channel is selected by employing a single activation of the image domain.
While this criterium is satisfactory for images of well-localized objects such as faces,  for instance, it turns to be not necessary the best alternative available for images with features distributed on the entire domain. Specifically, a single strong activation on a feature map can wrongly identify a channel as possessing the most representative features. To assess whether the selected channel is actually the most representative one, we performed several realizations of the feature extraction process. Our results suggest  that  the learned features at individual layers can be sensitive to the initial values of the weights, as  it shown on the second column of Fig.~\ref{fig:robustFailure}. This figure shows features maps that vary considerably when using different initializations of the weights. Compare this figure with the feature maps in Fig.~\ref{fig:inverseProb}. This lack of statistical robustness prompts us to we propose a more robust approach  in section \ref{deepeigenfeatures}.
%}

%

%
%%%%%%%%%%%%%%%%%%%%%%%%%%%%%%%%%%%%%%%%%%%%%%%%%%%%%%%
\clearpage
\newpage
 %***************************************************
 \subsection{Deep learning eigenfeatures} \label{deepeigenfeatures} 
  %***************************************************
%Using the $\alpha$ channel with the strongest activation as  criterion to select the most representative features of each sample is not necessary the best alternative available.
%Although this method highlights differences among the samples, the learned features at individual layers can be sensitive to the initial values of the weights in different realizations of the image classification process.
%While using a data set that is large enough would help to alleviate this weakness,  we propose a more robust approach.
%
In this section, instead of using the $\alpha$ channel, we consider simultaneously all  feature maps in a convolutional layer.
To discover the most representative features hidden in the feature maps, one idea is to assemble a matrix containing all the feature maps and then extract  the most representative features of this matrix.
SVD provides a convenient way to organize the feature maps into hierarchically ordered  contributions.
We briefly describe the SVD method in the next section.

%%%******************
\subsubsection{Singular value decomposition} \label{secSVD}
SVD is a  machine learning tool that has extraordinary applications~\cite{Hastie2001,Kutz2013,Alter2000,Kang2013,Fioranelli2015}. 
We are interested in analzying a data set $\mathbf{X} \in \mathbb{R} ^{n\times m}$.
 %
%\[
\begin{equation}
\mathbf{X}=
\left[
  \begin{array}{cccc}
    \vert & \vert &        & \vert \\
    {X}_{1}   & {X}_{2}   & \ldots & {X}_{m}   \\
    \vert & \vert &        & \vert 
  \end{array}
\right]
 \label{eqn:matrixResponses}
\end{equation}
%\]
%
In our case, the columns  ${X_k} \in \mathbb{R} ^{n}$ are individual feature maps at a layer of the CNN. We consider the activations at ReLU~2.
The size of the response of this layer is resized to match the input image or $266\times 266$ and thus we arrange each of these responses into column vectors with $n=70756$ elements.
Each feature map has index $i = 1,2, ..., m$, with $m$ being the total number of feature maps.
In our case $m=128$ because the second convolutional layer of the CNN has $128$ channels.

The SVD is a unique matrix decomposition  that exist for every matrix  $\mathbf{X}$:
\begin{equation}
 \mathbf{X}=U \Sigma  \mathcal{V}^{\top}
 \label{eqn:svd}
\end{equation}
with $U$  and $\mathcal{V}$ being the left and right singular vectors, respectively. The symbol ${\top}$ indicates transpose.
 The details of SVD can be found in the literature~\cite{Strang1980}. For the purpose of this work, suffice it to say that SVD  is a matrix factorization into unitary matrices $U$  and $\mathcal{V}$ with orthonormal columns that are ordered hierarchically according to their importance,
 and $\Sigma$ is a diagonal matrix containing the singular values. 
 
 A convenient statistical interpretation of the SVD involves the correlation matrix $\mathbf{X}^{\top} \mathbf{X}$ defined as
\begin{equation}
\mathbf{X}^{\top} \mathbf{X}=
\left[
  \begin{array}{cccc}
    {X}_{1}^\top  {X}_{1}    &   {X}_{1}^\top  {X}_{2}  & \ldots  & {X}_{1}^\top  {X}_{m} \\
    {X}_{2}^\top  {X}_{1}    &   {X}_{2}^\top  {X}_{2}  & \ldots  & {X}_{2}^\top  {X}_{m} \\
    \ldots                                  &    \ldots                              & \ldots  &  \ldots   \\
    {X}_{m}^\top  {X}_{1}    &   {X}_{m}^\top  {X}_{2}  & \ldots  & {X}_{m}^\top  {X}_{m} \\

  \end{array}
\right]
\label{eqn:xtransx}
\end{equation}
where each entry ${X}_{i}^\top  {X}_{j}=\langle X_i, X_j\rangle$ represents the inner product between columns $i$ and $j$. In other words,  ${X}_{i}^\top  {X}_{j}$ accounts for the overlapping  between all pairs of columns. This matrix accounts for the correlation between all of the feature maps in  $\mathbf{X}$.

To bring the SVD into play, notice that Eqn.~\ref{eqn:svd} allows us to write $\mathbf{X}^{\top} \mathbf{X}=\mathcal{V} \Sigma^2 \mathcal{V}^{\top}$. Similarly, $\mathbf{X} \mathbf{X}^{\top}=U \Sigma^2 U^{\top}$. These expressions can be written as the following eigenvalue problems:
\begin{equation}
 \begin{array}{c}
 \mathbf{X}^{\top} \mathbf{X} \mathcal{V}=\mathcal{V} \Sigma^2 \\
 \mathbf{X} \mathbf{X}^{\top} U=U \Sigma^2
   \end{array}
   \label{eqn:eigenvalProb}
\end{equation}

It is clear from Eqn.~\ref{eqn:eigenvalProb} that the columns of $\mathcal{V}$ and $U$ are eigenvectors of  the correlation matrices $ \mathbf{X}^{\top} \mathbf{X}$ and $\mathbf{X} \mathbf{X}^{\top}$, respectively. Loosely speaking, since  the columns in $U$ are ordered according to their importance, then the data in  $ \mathbf{X}$  has its larger correlation  along the $u_1$ eigendirection. Where $u_1$ is the first column of the left singular eigenvector $U$.
These eigenvectors define directions along which all the feature maps in $\mathbf{X}$ have the largest variance. In this context, the eigenvectors are the principal component directions  PC$_i$.
%
%++++++++++++++++++++++++++++++++++++++++++++++++++++++
%
\subsubsection{Eigenvectors of the feature maps}
The process of feature discovery of a test image is summarized in Fig.~\ref{fig:cnnScheme}.
To obtain the most representative features of a test image of the absolute gradient of a sample of material, firstly we  feed the test image  into an already trained CNN with the  architecture defined in Table 2.
Secondly, we collect  all the channels containing the feature maps of the test image and use each feature map as a column vector $X_i$  for the matrix $\mathbf{X}$ in Eqn.~\ref{eqn:matrixResponses}. 
We then apply the SVD method to the whole library of feature maps obtained at ReLU~2, which follows the second convolutional layer.
%
%\textcolor{red}{}
The feature maps can be projected onto the subspace spanned by the eigenvectors to obtain the weights of the linear combination of eigenvectors needed to reconstruct them. 
The fast decay of the singular values shown in Fig.~\ref{fig:cnnScheme}d suggests that the first eigenvector, $u_1$, can be used to describe the most relevant features  at the second convolutional layer.
The eigenvector $u_1$ represents the direction in which $\mathbf{X}$ has the largest variance.
%
%The feature map with the largest projection onto $u_1$ is the most representative feature map and 
%As an illustration, the one-term approximation to any feature map in  $\mathbf{X}$ is $X_k=\sigma_1 u_1 v_1^\top  i_k$, where $u_1$ and $v_1$ are the left and right singular vectors, respectively, and  $i_k$ is a label vector for the $k$-st feature map.
% This is the best rank-1 approximation to any feature map in  $\mathbf{X}$.
%
%
In terms of visual representation, the bright regions of $u_1$ exhibit the greatest variance and, therefore,  contain the most representative characteristics of the entire library of  feature maps at a given layer. The eigenvector $u_1$ is  chosen because it represents the most significant components of a test image.
Notice that since $u_1$ is computed from the features map, this quantity represents qualitatively the content of TV on the images.
%

% \begin{figure}[h]
%\begin{center}
%\includegraphics[width=70mm]{./figures/singularvaluesT3.pdf}    
%\end{center}
%\caption{
%Singular values obtained from the singular value decomposition of  the matrix $\mathbf{X}$ containing 128 channels of feature maps.
%The channels are the response to the classification process of a typical image of sample No. 3.
%}
%\label{fig:sa}
%\end{figure}
 
\begin{figure}[h]
\begin{center}
\includegraphics[width=100mm]{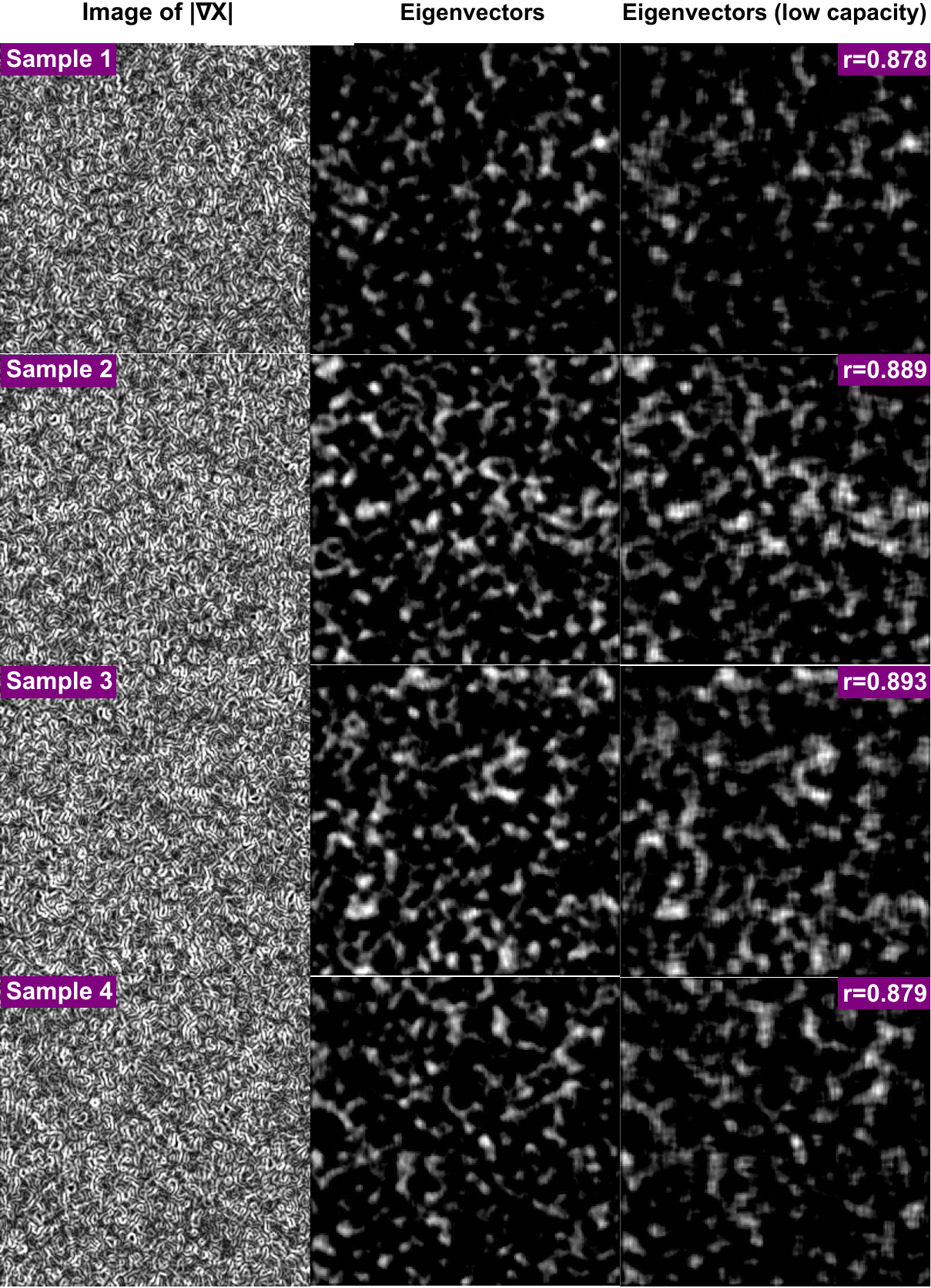} 
\end{center}
\caption{
Features of the test images of samples 1-4 (top to bottom) obtained using the eigenvectors of the library of response maps at  ReLU~2 as described by Eq.~\eqref{eqn:matrixResponses}.
The eigenvectors highlight differences among samples.
First column shows test gradient images, $X=|{\nabla Y}|$.
Second and third column show features of the images of the first column obtained using eigenvectors of responses of
(i) the CNN described in Table 1 and (ii) the CNN of reduced capacity as described in the text, respectively.
The correlation coefficient, $r$, between the second and third columns is presented for comparison.
Notice that the features in sample  No. 3 are notably more prominent than those in the other samples, as expected.
}
\label{fig:inverseProbEigen}
\end{figure}

We apply the above process to extract the  deep learning eigenvectors of typical images of samples 1-4 and the result is shown in the second column of Fig.~\ref{fig:inverseProbEigen}.
Notice that the eigenvectors are similar in appearance to the $\alpha$ channel shown in Fig.~\ref{fig:inverseProb} from the previous section. 
 The  deep learning eigenvectors posses the notable difference of being   statistically  robust in the sense that they seem to be less sensitive to the initial values of the weights, which is a desired property. 
 
\begin{figure}[h]
\begin{center}
\includegraphics[width=100mm]{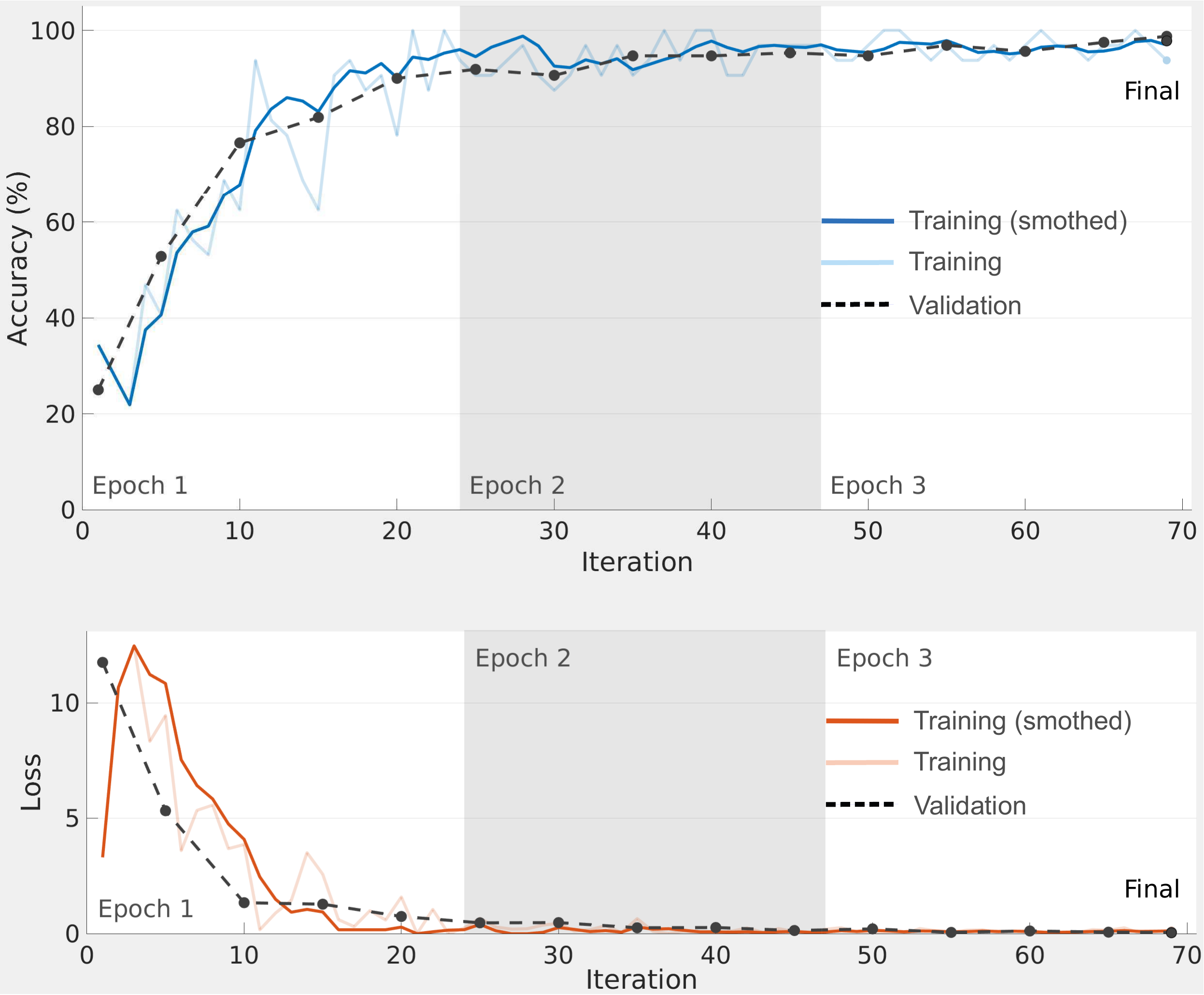}    
\end{center}
\caption{
Training progress of the CNN with reduced capacity using the hyperparameters described in the text.
Using a mini-batch size of 32, 
it takes 23 iterations to all of the training samples pass through the learning algorithm.
(a) Training and validation accuracy reaches more or less  the same high values after 70 iterations. 
(b) The steady decrease of the training and validation loss functions suggests that there is not significative overfitting. 
}
\label{fig:trainingSmall}
\end{figure}

It is crucial to verify that the eigenvectors associated to the samples are  correct and robust.
This is specially necessary because of the limited size of our database, although we have employed augmentation to enlarge the database.
To this end, we have employed several approaches, namely weight regularization and addition of dropout to the fully-connected layer. 
Another technique to make sure that overfitting is not significative consists in reducing  the capacity of the network~\cite{Chollet2018}.
One way to implement this in practice is by comparing models possessing different number of hidden units~\cite{Bishop1995}.
We assessed this idea in our model by removing the last convolutional block and additionally we substantially decreased the number of learnable parameters by halving the number of channels in each one of the first two layers, thereby leaving the first two convolutional layers with 64 channels each.
Figure~\ref{fig:trainingSmall}  shows the training progress of the simpler model with reduced capacity, in which we have used a mini-batch of size 32. The steady decrease of the training and validation loss, suggests that there is no significative overfitting and the final accuracy is similar to that of the original model.
We observed no significative reduction in classification accuracy and more importantly, the eigenvectors in samples 1-4 remain more or less unchanged, as show in the third column of Fig.~\ref{fig:inverseProbEigen}. 
A strong correlation coefficient close to $r=0.9$ between the features obtained using the  full model and the model with reduced capacity indicates that the simpler network model captures very well the most representative features for classification.

 %%%************************************

\section{Discussion and conclusions}
A highly accurate CNN classifies  images of the magnitude of the gradient of X-ray of samples of epoxy resins.
%new
In addition to the epoxy resins used in this work, other types of resins can be studied provided that the samples contain inhomogeneities in the spatial density that are larger than the resolution of the apparatus.
%ok
The feature maps of the intermediate layers of the CNN contain the most representative low-level features in a test image.
The strongest activated channel produces an image of these features  providing us with a simple visual summary of a given sample.
However different realizations of the classification process result in slightly different features. For images with  well-defined segments, such as faces or objects, this method can be considered adequate. In the case of  X-ray images of resins, the descriptive features are distributed throughout the whole domain and therefore a better visualization of the representative features is desirable.

SVD provides a means to expand a matrix of feature maps, $\mathbf{X}$, in terms ordered hierarchically as:
\begin{equation}
\mathbf{X}= \sigma_1 u_1 v_1^{\top}+ \sigma_2 u_2 v_2^{\top}+ \ldots + \sigma_m u_m v_m^{\top}
   \label{eqn:svdExpan}
\end{equation}

Truncating the sum in  Eqn.~\ref{eqn:svdExpan}  to include only the fist $r$ terms results in the matrix
$\widetilde{\mathbf{X}}=\widetilde{U} \widetilde{\Sigma}  \widetilde{\mathcal{V}}^{\top}$.
The Eckard-Young theorem~\cite{Eckart1936} guarantees that the best  approximation to  $\mathbf{X}$ of rank-r is $\widetilde{\mathbf{X}}$, according to
\begin{equation}
% \| \mathbf{X} - \widetilde{\mathbf{X}}\| 
\underset{ \widetilde{\mathbf{X}} \text{ s.t. rank}(\mathbf{X})=r }{\arg\min} \| \mathbf{X} - \widetilde{\mathbf{X}}\|_F=
\widetilde{U} \widetilde{\Sigma}  \widetilde{\mathcal{V}}^{\top}
    \label{eqn:eckard}
\end{equation}
Where $ \| {\cdot}\|_F$ is the Frobenius norm. Then the best approximation to $\mathbf{X}$ that has rank $r=1$ is given by
\begin{equation}
\widetilde{\mathbf{X}}=  \sigma_1 u_1 v_1^{\top}
   \label{eqn:svdExpanRank1}
\end{equation}
Where $u_1$ is the first column in the left singular eigenvector which has the same size as the feature maps $X_i$. Therefore, the eigenvector $u_1$ is an optimal representation of the rank-1 truncation of the whole library of feature maps and contains the most representative  features of the original test image of a given sample.
The one-term approximation to any feature map in  $\mathbf{X}$ is written as  $X_k=\sigma_1 u_1 v_1^\top  i_k$, where  $i_k$ is a label vector for the $k$-st feature map.

Some advantages of proposed approach are the following:
(i) The eigenvectors of the feature maps  are in agreement with the results obtained using the strongest activated channel. This means that both methods highlight approximately the same region in the domain of  the test image, although the eigenvectors appear to have a much clear appearance than the features in the  strongest activativated channel.
(ii) The eigenvectors are statistically robust in the sense that they remain unchanged when retraining the CNN with different initial values of the weights.
(iii) More importantly, the eigenvectors seem to be   statistically robust  across different network architectures. To demonstrate this aspect, we  retrained the AlexNet~\cite{Krizhevsky2012}  to classify 4 samples of materials. We then used the SVD-based approach to summarize the feature maps and observed that the resulting eigenvectors have similar appearance to what is shown in Fig.~\ref{fig:inverseProbEigen}.

The SVD-based method to extract the most representative features of an X-ray image can be appropriate to a large variety of applications, including polymers, metal alloys, among others~\cite{Avalos}. Since opposite signs of $u_1$ represent the same eigenvector, a future work should  focus on developing an unsupervised selection of  the appropriate sign of the eigenvectors for feature identification.

%%%%%%%
%\appendix
%%%%%%%

\paragraph{Data availability}
 The raw/processed data required to reproduce these findings cannot be shared at this time as the data also forms part of an ongoing study.

\paragraph{Acknowledgment}
This work was partially supported by 
Cross-ministerial Strategic Innovation Promotion Program
(SIP), `Structural Materials for Innovation' and `Materials
Integration' for Revolutionary Design System of Structural
Materials,
and the support of KAKENHI Grants-in-Aid no.18H05482.

%%%REFERENCES%%%
\section*{References}
%\clearpage
%\bibliographystyle{unsrt}  %unsrt, ieeetr, unsrtnat
\bibliography{ML,xiesref} %.bib file

\end{document}